\documentclass[runningheads]{llncs}

\usepackage{eccv}

\usepackage{eccvabbrv}

\usepackage{graphicx}
\usepackage{booktabs}

\usepackage[accsupp]{axessibility}  %

\usepackage[most]{tcolorbox}

\definecolor{ipsamblue}{HTML}{2E6092}
\definecolor{ipsamlightblue}{HTML}{F5F8FC}

\newtcolorbox{showcaseabstract}{
    enhanced,
    breakable,
    colback=ipsamlightblue,
    colframe=ipsamblue!75,
    boxrule=0.7pt,
    arc=2.2mm,
    outer arc=2.2mm,
    left=3.2mm,
    right=3.2mm,
    top=2.8mm,
    bottom=2.8mm,
    before skip=8pt,
    after skip=11pt
}

\usepackage{hyperref}
\usepackage{pifont}
\usepackage{enumitem} 
\usepackage[misc]{ifsym}
\usepackage{orcidlink}
\usepackage{algorithm}
\usepackage{algpseudocode}
\algrenewcommand\algorithmiccomment[1]{\hfill{\scriptsize$\triangleright$~#1}}
\algrenewcommand\algorithmicrequire{\textbf{Input:}}
\algrenewcommand\algorithmicensure{\textbf{Output:}}
\usepackage{amsmath}
\usepackage{amssymb}
\usepackage{wrapfig}
\usepackage{subcaption}
\usepackage{colortbl}
\newcommand{\cmark}{\ding{51}}
\newcommand{\xmark}{\ding{55}}
\usepackage{xcolor}
\usepackage{arydshln}
\usepackage{multirow}
\definecolor{mygray}{gray}{.92}
\usepackage{pifont}

\newcommand{\figref}[1]{Fig. \ref{#1}}

\newcommand{\secref}[1]{$\S~$\ref{#1}}

\def\ie{\emph{i.e.}}
\def\eg{\emph{e.g.}}

\usepackage[misc]{ifsym}
\begin{document}
\title{IP-SAM: Rethinking Prompt-Conditioned Segmentation for Prompt-Absent Deployment}

\titlerunning{IP-SAM: Rethinking Prompt-Conditioned Segmentation}

\renewcommand\footnotemark{}
\author{Huiyao Zhang $^{1,2}$ \and
Jin Bai $^{1,2}$ \and  
Rui Guo $^{1,2}$ \and 
JianWen Tan $^{1,2}$ \and 
HongFei Wang $^{1,2}$ \and 
Ye Li $^2$ $^{(\textrm{\Letter})}$   \\
\thanks{$\textrm{\Letter}$~Corresponding authors.}
}

\authorrunning{H. Zhang et al.}

\institute{ $^1$University of Chinese Academy of Sciences, \\ $^2$Technology and Engineering Center for Space Utilization, Chinese Academy of Sciences \\
\email{zhanghuiyao25@csu.ac.cn, baijin25@mails.ucas.ac.cn, liye@csu.ac.cn}}

\maketitle
\begin{showcaseabstract}
\small

\noindent
{\bfseries\color{ipsamblue}Abstract}\quad
Prompt-conditioned foundation segmenters have emerged as a dominant
paradigm for image segmentation, where explicit spatial prompts
(e.g., points, boxes, masks) guide mask decoding. However, many
real-world deployments require fully automatic segmentation, creating
a structural mismatch: the decoder expects prompts that are unavailable
at inference. Existing adaptations typically modify intermediate
features, inadvertently bypassing the model's native prompt interface
and weakening prompt-conditioned decoding. We propose IP-SAM, which
revisits adaptation from a prompt-space perspective through prompt-space
conditioning. Specifically, a Self-Prompt Generator (SPG) distills image
context into complementary intrinsic prompts that serve as coarse regional
anchors. These cues are projected through SAM2's frozen prompt encoder,
restoring prompt-guided decoding without external intervention. To suppress
background-induced false positives, Prompt-Space Gating (PSG) leverages
the intrinsic background prompt as an asymmetric suppressive constraint
prior to decoding. Under a deterministic no-external-prompt protocol,
IP-SAM achieves state-of-the-art performance across four camouflaged
object detection benchmarks with only 21.26M trainable parameters.
Furthermore, the proposed conditioning strategy generalizes beyond COD
to medical polyp segmentation.

\par\medskip


\par\medskip




\end{showcaseabstract}


\section{Introduction}
\label{sec:intro}

Prompt-conditioned segmentation has emerged as a new paradigm for visual foundation models. Instead of directly predicting masks, these models rely on explicit spatial prompts—such as points, boxes, or masks—to guide the decoding process. Representative systems such as the Segment Anything Model (SAM) and its successor SAM2~\cite{kirillov2023segment, ravi2024sam} demonstrate remarkable flexibility under interactive settings, enabling users to steer segmentation through simple spatial cues.

However, this paradigm reveals a fundamental mismatch in fully automatic deployment. In many real-world scenarios, segmentation must operate without human interaction, leaving the model strictly prompt-absent at inference time. This creates a structural paradox: the decoder is designed to rely on prompt-conditioned signals, yet no prompts are available during deployment.

This issue becomes particularly evident in Camouflaged Object Detection (COD), where targets blend into surrounding textures~\cite{fan2020camouflaged}. Without disambiguating prompts, models frequently activate visually similar background regions, leading to unstable masks and severe background leakage. Although we evaluate on COD as a challenging stress-test benchmark, the prompt-absent deployment problem arises broadly across automatic segmentation tasks.

To address this challenge, most existing adaptations modify intermediate features by introducing adapters, lateral fusion modules, or task-specific decoders~\cite{chen2023sam, gao2024multi}. While these feature-space modifications improve task alignment, they inadvertently bypass the native prompt interface that the architecture was originally designed to exploit. As a result, the prompt-conditioned decoding pathway remains weakly activated, leading to suboptimal adaptation. We argue that bypassing the prompt interface fundamentally breaks the design assumption of prompt-conditioned segmenters.

To resolve this mismatch, we propose Intrinsic Prompting SAM (IP-SAM), which revisits adaptation from a prompt-space perspective through explicit prompt-space conditioning. Instead of injecting task information into intermediate features, IP-SAM synthesizes intrinsic prompts that activate the model through its native prompt interface. Concretely, a Self-Prompt Generator (SPG) distills aliased image contexts into complementary prompt candidates that serve as coarse regional anchors. These cues are projected through SAM2's frozen prompt encoder, translating task-specific signals into the model's native prompt manifold and restoring prompt-conditioned decoding.

Furthermore, camouflaged scenes exhibit severe foreground–background ambiguity. Traditional feature fusion tends to symmetrically mix foreground and background cues, often amplifying deceptive textures. To address this, we introduce Prompt-Space Gating (PSG), which uses the intrinsic background prompt as an asymmetric suppressive constraint. By filtering deceptive activations prior to decoding, PSG neutralizes background-induced false positives through prompt-space suppression.

We evaluate IP-SAM under a deterministic no-external-prompt protocol across four COD benchmarks. With only 21.26M tunable parameters—optimizing SPG, PSG, and a task-specific mask decoder alongside image-encoder LoRA while keeping the prompt encoder strictly frozen—IP-SAM consistently outperforms heavy specialist COD models and recent SAM adaptation baselines. In essence, IP-SAM restores the missing prompt signal by synthesizing intrinsic prompts and conditioning SAM2 through its native prompt interface.

\noindent \textbf{Contributions.}
(i) We identify the prompt-absent deployment paradox in prompt-conditioned segmentation models and introduce a prompt-space conditioning paradigm that restores the native prompt interface without external prompts.  
(ii) We propose Prompt-Space Gating (PSG), which leverages intrinsic background prompts as asymmetric suppressive constraints to neutralize background-induced false positives prior to decoding.  
(iii) With only 21.26M trainable parameters, IP-SAM achieves state-of-the-art performance on COD benchmarks and demonstrates strong zero-shot cross-dataset transfer to medical polyp segmentation.

\section{Related Work}
\label{sec:related}

\subsection{Camouflaged Object Detection}
Camouflaged Object Detection (COD) aims to segment objects intentionally concealed within their surroundings, a task plagued by extreme foreground-background aliasing~\cite{fan2020camouflaged, mei2021camouflaged}. While early approaches relied on heavy task-specific designs (e.g., explicit boundary modeling, multi-scale aggregation)~\cite{fan2020camouflaged, li2021uncertainty, zhang2020uc}, recent Transformer-based models~\cite{pang2022zoom, ji2023deep, yin2024camoformer} have improved global structural modeling. Nevertheless, specialist COD models trained from scratch on limited datasets fundamentally struggle with domain generalization and diverse camouflage topologies. This limitation logically motivates the utilization of large-scale pretrained foundation segmenters, which inherently possess richer semantic representations and geometric priors.

\subsection{Adapting SAM for Downstream Tasks}
The Segment Anything Model (SAM)~\cite{kirillov2023segment, ravi2024sam} has inspired extensive domain adaptation efforts~\cite{ma2024segment, chen2023sam}. To handle prompt-absent scenarios, prevalent methods rely heavily on feature-space adaptation, injecting structural patches---such as adapters or heavy decoding heads---into the backbone or decoder~\cite{chen2023sam, chen2024sam2, gao2024multi}. Recent SAM-based COD models largely follow this feature-centric paradigm~\cite{gao2024multi, chen2024sam, liu2025improving, yuan2025sam2}. However, by predominantly modulating intermediate features, these methods often underutilize SAM's defining characteristic: its prompt-conditioned decoding interface. This raises a critical question: rather than brute-forcing intermediate features, how can we translate task requirements directly into SAM's native prompt manifold while preserving prompt-conditioned decoding?

\subsection{Automatic Prompting and Parameter-Efficient Tuning}
To automate SAM, recent works train auxiliary networks to generate explicit geometric cues (points, boxes)~\cite{shaharabany2023autosam, chen2024rsprompter, xie2024pa}, explore prompt-free deployment via self-prompting and test-time calibration~\cite{zhang2025hierarchical, chen2025aop}, or leverage class activation maps and domain-adaptive prototypes for medical segmentation~\cite{wei2024prompting}. In the context of COD, however, automatically generated explicit prompts frequently collapse. Due to extreme visual aliasing, predicted boxes easily misregister on deceptive distractors, directly injecting severe background-induced false positives into the decoder. Furthermore, while some prior arts generate dense conceptual prompts~\cite{liu2022prompt}, they typically rely on symmetric fusion without actively isolating background interference.

Our approach fundamentally avoids this fragility. Rather than relying on discrete coordinate predictions, heavy feature-space addition, or test-time optimization, we explicitly translate complementary dense spatial priors into SAM's frozen prompt manifold. Crucially, we distinguish our paradigm from prior self-prompting routes by introducing Prompt-Space Gating (PSG) to preemptively filter noise via asymmetric suppression. Resolving semantic conflicts prior to decoding maximizes the efficacy of Parameter-Efficient Fine-Tuning (PEFT) like LoRA~\cite{hu2022lora}. Existing methods often force PEFT to shoulder the entire feature realignment burden without reliable prompts, which is inefficient. By reactivating the proper prompt manifold, IP-SAM achieves exceptional robustness with efficient updates.

\section{Method}
\label{sec:method}

\subsection{Overview}
\begin{figure*}[tb]
\centering
\includegraphics[width=1\textwidth]{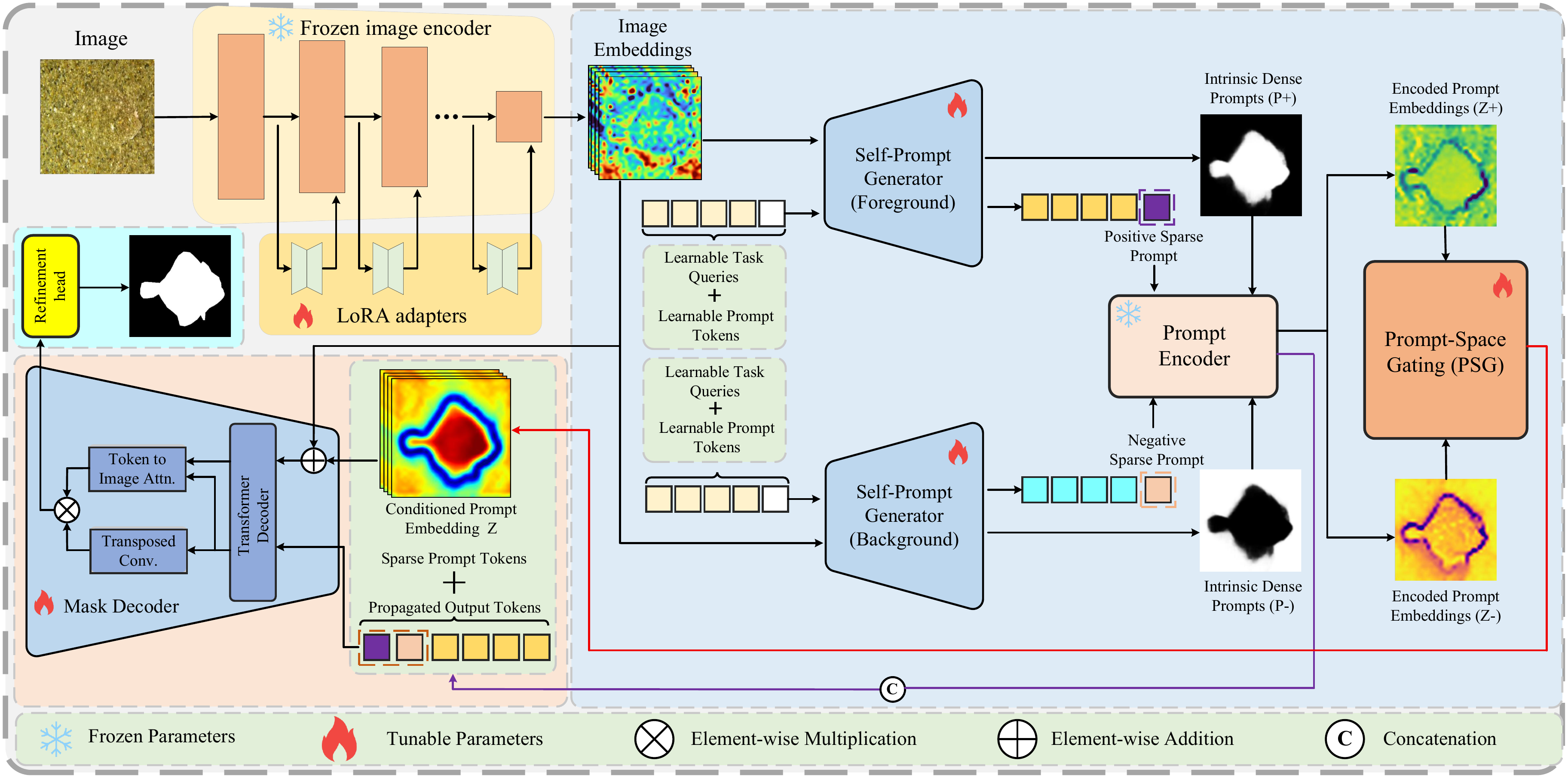}
\caption{\textbf{Overall architecture of IP-SAM.} Shifting the paradigm to prompt-space conditioning, IP-SAM utilizes a Self-Prompt Generator (SPG) to synthesize complementary intrinsic prompts ($P^\pm$). These are projected into SAM2's native manifold via the frozen prompt encoder. Prior to decoding, Prompt-Space Gating (PSG) leverages the negative embedding ($Z^-$) to asymmetrically suppress deceptive cues in the positive embedding ($Z^+$). The purified condition ($Z$) explicitly steers the mask decoder. For visual clarity, the optional auxiliary structural regularizer (ablated in \secref{sec:ablation}) is omitted.}
\label{fig:arch}
\end{figure*}

\begin{figure*}[tb]
\centering
\includegraphics[width=1\textwidth]{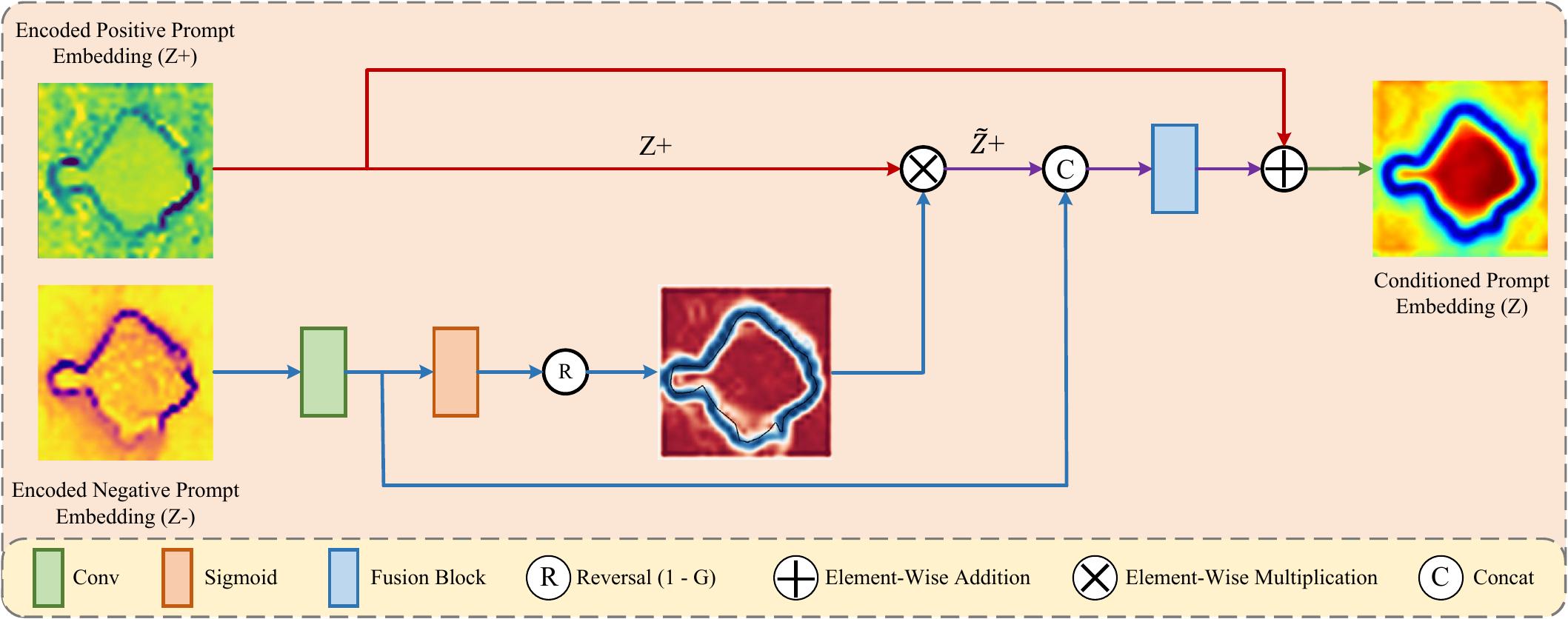}
\caption{\textbf{Detailed schematic of Prompt-Space Gating (PSG).} PSG isolates and eliminates background-induced false positives before they enter the decoder. It formulates a localized gate decision map from the negative prompt embedding ($Z^-$) to physically filter deceptive cues from the positive counterpart ($Z^+$). The suppressed feature ($\widetilde{Z}^+$) is then fused with unactivated background cues and compensated via an anchored residual connection to yield the robust Conditioned Prompt Embedding ($Z$).}
\label{fig:psg_detail}
\end{figure*}

To resolve the structural paradox of prompt-absent deployment, IP-SAM revisits adaptation by conditioning SAM2 through its native prompt space. Specifically, IP-SAM synthesizes task-aware prompts and feeds them through SAM2's native prompt interface. As illustrated in \figref{fig:arch}, a Self-Prompt Generator (SPG) synthesizes dense intrinsic prompts. These cues are projected through SAM2's frozen prompt encoder, structurally aligning the task with the prompt-conditioned decoding process used by our task-specific decoder. To eradicate false positives, Prompt-Space Gating (PSG) (\figref{fig:psg_detail}) imposes background-derived asymmetric suppressive constraints strictly prior to decoding.

\subsection{Intrinsic Prompt Generation and Frozen Encoding}
\noindent \textbf{Image embedding.}
Given an input image $I$, the SAM2 image encoder $\mathcal{F}$ extracts high-dimensional visual semantics to produce an image embedding $E = \mathcal{F}(I) \in \mathbb{R}^{C \times H \times W}$.

\noindent \textbf{Self-Prompt Generator (SPG).}
Instead of relying on heuristic pseudo-prompts (boxes/points) that are brittle under extreme camouflage, we learn dense mask-like priors as intrinsic prompts. We design a dual-branch Self-Prompt Generator $\mathcal{G}$ driven by learnable task queries $Q \in \mathbb{R}^{N \times d}$. Specifically, $E \in \mathbb{R}^{256 \times 32 \times 32}$ denotes the highest-semantic-level feature map produced by the FPN neck, at $\frac{1}{16}$ of the input resolution. We flatten $E$ into a sequence of length $(HW) \times C$ before cross-attention. $\mathcal{G}$ employs a 2-layer TwoWayTransformer decoder (structurally identical to SAM's default transformer block, with $d=256$, 8 heads, MLP dim 2048). Here, $Q$ (comprising standard sparse prompt slots $K$ and $K_m=4$ learnable mask tokens) extracts complementary foreground and background semantics from the flattened $E$ via cross-attention (\ie, $\mathrm{Attn}(Q, E, E)$). Through a dedicated projection head (which upsamples the feature map by a factor of 4 using bilinear upsampling and a $3\times3$ convolution), SPG distills this aggregated context into complementary dense prompt logits ($P^+, P^- \in \mathbb{R}^{1 \times 128 \times 128}$) alongside continuous sparse prompt tokens ($S^+, S^-$). Notably, the positive branch additionally yields a set of specialized propagated output tokens $\mathcal{T}_{\text{prop}}$ (depicted as the purple block in \figref{fig:arch}) to serve as strong semantic anchors:
\begin{equation}
\{P^+, S^+, \mathcal{T}_{\text{prop}}\}, \{P^-, S^-\} = \mathcal{G}(E, Q),
\label{eq:spg}
\end{equation}
where $\mathcal{T}_{\text{prop}} \in \mathbb{R}^{K_m \times C}$ strictly matches the number ($K_m=4$) and dimensionality ($C=256$) of SAM2's default mask tokens. Given the extreme foreground-background aliasing in COD, these synthesized prompts inherently inherit positional ambiguity and structural noise from the camouflaged scene. They provide essential high-level spatial intents, while the mask decoder performs fine-grained disambiguation.

\noindent \textbf{Manifold projection and sparse alignment.}
We feed the real-valued dense logits $P^\pm$ directly into SAM2's frozen native prompt encoder $\mathcal{E}$, yielding the encoded dense prompt embeddings $Z^\pm = \mathcal{E}(P^\pm) \in \mathbb{R}^{C \times H_e \times W_e}$. In practice, although native masks are binary, we intentionally feed continuous logits into the frozen prompt encoder without explicit binarization. Empirically, we found that binarizing the masks or applying temperature scaling consistently degrades performance, suggesting that continuous logits provide richer spatial uncertainty for prompt conditioning while remaining compatible with the encoder’s embedding space. For sparse prompts, SPG bypasses the discrete coordinate-to-embedding step by directly outputting continuous token embeddings $S^\pm$. We use the same number of sparse prompt slots as SAM2 (denoted $K$) and add the corresponding pre-trained (positive/negative) point-type embeddings to form the encoded sparse conditions $U^\pm \in \mathbb{R}^{K \times C}$. Keeping $\mathcal{E}$ strictly frozen translates task-specific cues into SAM2's native prompt embedding space used for decoding, formalizing our prompt-space conditioning.

\subsection{Prompt-Space Gating (PSG)}
In camouflaged scenes, standard symmetric feature fusion often amplifies deceptive textures. Although the prompt encoder can successfully embed the negative dense prior $P^-$ into the native prompt space, directly providing the resulting dense background embedding to the decoder as a symmetric condition tends to amplify background activations under camouflage, leading to more false positives. Therefore, we use the negative dense branch exclusively to form a suppressive gate prior to decoder interaction.

First, the negative dense prompt embedding $Z^-$ is processed by a predictor $\phi$ (a $3\times3$ conv reducing 256 to 64 channels, GELU, and a $1\times1$ projection to 1 channel). We formulate a localized gate decision map by applying a sigmoid activation $\sigma$ and using its complement to suppress background-correlated activations:
\begin{equation}
G = \sigma(\phi(Z^-)), \quad \widetilde{Z}^{+} = Z^{+} \odot (1-G).
\label{eq:psg_gate}
\end{equation}
Here, the complement $(1-G)$ serves as a suppressive gate that neutralizes highly correlated deceptive elements within the positive dense prompt $Z^+$. This yields the purified target feature $\widetilde{Z}^{+}$.

To restore structural integrity while effectively filtering false positives, we process the channel-wise concatenated features $[\widetilde{Z}^{+},\, \phi_{\text{feat}}(Z^-)]$ through a fusion block $\psi$ (a $3\times3$ conv from 512 to 256 channels, LayerNorm, GELU, and a $1\times1$ conv maintaining 256 channels). We perform an anchored residual correction ($\oplus$) with the original positive prompt embedding: $Z = \psi([\widetilde{Z}^{+},\, \phi_{\text{feat}}(Z^-)]) + Z^+$. Crucially, the residual is anchored to the original $Z^+$ rather than the suppressed $\widetilde{Z}^+$. This is because the asymmetric gate may inevitably attenuate some fine-grained foreground signals during suppression; anchoring to $Z^+$ allows $\psi$ to specialize in predicting the isolated noise component (a dynamic noise cancellation signal), rather than simultaneously reconstructing the damaged foreground and canceling noise.

\subsection{Decoding and Optional Enhancements}
\noindent \textbf{Prompt-conditioned decoding.}
To maintain compatibility with SAM2's prompt interface, we provide our task-specific mask decoder with a comprehensive sparse condition. As depicted in \figref{fig:arch}, we concatenate the continuous sparse prompt tokens from both branches to form a unified sparse condition $[U^+; U^-]$. Furthermore, the SAM2 decoder inherently relies on a set of standard learnable output tokens (an IoU token and multiple mask tokens). To fully leverage our SPG, we replace the default mask tokens with the propagated output tokens $\mathcal{T}_{\text{prop}}$ from the positive SPG branch, while retaining the standard IoU token (denoted as $T_{\text{iou}}$). This maintains the expected token sequence length while injecting robust semantic anchors. The task-specific mask decoder $\mathcal{D}$ is thus driven by the image embedding $E$, the condition-purified dense prompt embedding $Z$, the concatenated sparse tokens $[U^+; U^-]$, and the customized output tokens $[T_{\text{iou}}; \mathcal{T}_{\text{prop}}]$, yielding a robust coarse mask:
\begin{equation}
M_c = \mathcal{D}\big(E, Z, [U^+; U^-], [T_{\text{iou}}; \mathcal{T}_{\text{prop}}]\big).
\end{equation}
This achieves robust decoding under camouflage while preserving the native conditioning interface.

\noindent \textbf{Optional enhancements.}
Optionally, we add two lightweight components (ablated in \secref{sec:ablation}): (i) an auxiliary structural regularizer using the detached coarse mask $M_c$ to form a soft gate (\eg, $\sigma(M_c)$) suppressing multi-scale FPN features; and (ii) a pixel refinement head $\mathcal{R}$ (fusing the coarse mask, dense prompt, and 256-channel image features via a 64-channel bottleneck) to predict a residual correction $M = M_c + \mathcal{R}(\cdot)$ via a zero-initialized $1\times1$ conv for ultra-fine boundary alignment.

\subsection{Training Objective and Efficient Adaptation}

\noindent \textbf{Losses \& Adaptation.}
Let $Y \in \{0,1\}^{1 \times H_0 \times W_0}$ be the ground truth. We supervise SPG with complementary targets: $\mathcal{L}_{\text{SPG}} = \mathrm{BCE}(\sigma(P^+), Y) + \mathrm{BCE}(\sigma(P^-), 1-Y)$. Although SPG is supervised to learn mask-like priors, the final prediction is produced by a task-specific mask decoder conditioned on their prompt-space embeddings, rather than directly from $P^\pm$. The dense mask loss $\mathcal{L}_{\text{mask}}$ combines standard BCE, IoU, and L1 components. The overall objective is $\mathcal{L} = \lambda_{\text{spg}}\mathcal{L}_{\text{SPG}} + \lambda_{\text{c}}\mathcal{L}_{\text{mask}}(M_c, Y) + \lambda_{\text{r}}\mathcal{L}_{\text{mask}}(M, Y)$, where $\lambda_{\text{r}}=0$ when refinement is disabled. We utilize LoRA on the $qkv$ and output projections of the SAM2 image encoder across all Transformer blocks to enable lightweight domain adaptation. Importantly, the prompt encoder $\mathcal{E}$ remains strictly frozen to preserve the pre-trained prompt embedding manifold. A lightweight task-specific mask decoder is implemented following the architectural design of SAM's mask decoder, but is trained entirely from scratch without using SAM2 pretrained weights. Overall, this results in only 21.26M trainable parameters.

\section{Experiments}
\label{sec:exp}

\subsection{Experimental Settings}
\label{sec:exp_setting}

\noindent \textbf{Datasets and metrics.}
We evaluate on four standard COD benchmarks: CAMO~\cite{le2019anabranch}, CHAMELEON~\cite{skurowski2018animal}, COD10K~\cite{fan2020camouflaged}, and NC4K~\cite{lv2021simultaneously}.
Following~\cite{he2023weakly,he2023weakly2,hu2025int,hu2024relax,hu2024leveraging}, we train on the COD10K and CAMO training splits, and test on their respective test sets plus the full CHAMELEON and NC4K datasets. Performance is assessed using Mean Absolute Error (MAE) $\downarrow$, weighted F-measure ($F^{\omega}$) $\uparrow$, Structure-measure ($S_m$) $\uparrow$, and Enhanced-alignment measure ($E_{\phi}$) $\uparrow$.

\noindent \textbf{Implementation \& evaluation.}
IP-SAM is built upon the SAM2-L (Hiera-Large) image encoder. Images are resized to $512 \times 512$. We train end-to-end for 100 epochs using AdamW~\cite{loshchilov2017decoupled} (LR $1 \times 10^{-4}$, batch size 1) on a single RTX 3090. Loss weights are fixed at $\lambda_{\text{spg}}\!=\!\lambda_{\text{c}}\!=\!\lambda_{\text{r}}\!=\!1.0$. To ensure absolute fairness, all baselines (specialist and SAM-based) are explicitly re-trained under the identical data and resolution. During evaluation, we strictly adhere to a deterministic prompt-absent protocol: standard SAM/SAM2 baselines use empty tokens and zero-mask inputs, whereas IP-SAM synthesizes intrinsic conditions ($U^\pm, Z^\pm$). For SAM2-AMG, we select the top-1 mask via its native \texttt{predicted\_iou}, since top-$k$ selection requires ground-truth knowledge.

\noindent \textbf{Parameter-efficient tuning.}
To isolate the effectiveness of our prompt-space conditioning, we keep the native prompt encoder strictly frozen. Param and T-Param denote the total and trainable parameters of the instantiated inference model, respectively, reported in millions (M). Our tunable parameters consist solely of the lightweight SPG/PSG modules, our task-specific mask decoder, and a LoRA (rank $r=32$) applied to the attention projections of the SAM2 image encoder.

\subsection{Comparisons with State-of-the-Art Methods}
\label{sec:sota}

\noindent \textbf{Quantitative superiority.}
Table~\ref{tab:sota} shows IP-SAM establishes a new state-of-the-art in the prompt-absent setting. On the heavily cluttered CAMO dataset, IP-SAM pushes MAE to 0.032 and $S_m$ to 0.912, outperforming both the strongest specialist (ZoomNeXt, $S_m$ 0.892) and the feature-adapted baseline (SAM2-Adapter, $S_m$ 0.884). Furthermore, IP-SAM substantially outperforms SAM2-AMG. On CHAM\allowbreak-ELEON, SAM2-AMG trails the naive null-prompt baseline (MAE 0.500 vs. 0.182); lacking confidence on concealed targets, AMG's top-1 proposal frequently gravitates toward well-defined background regions, drastically inflating errors.

\begin{table}[tb] 
    \centering
    \caption{\textbf{Quantitative comparison (no-external-prompt).} Best results are boldfaced and IP-SAM is highlighted in gray.}
    \label{tab:sota}
    \setlength{\tabcolsep}{1.5pt} 
    \renewcommand{\arraystretch}{1} 
    \scriptsize
    \resizebox{\linewidth}{!}{
    \begin{tabular}{l c c c c c c c c c c c c c c c c c c}
        \toprule
        \multirow{2}{*}{\textbf{Method}} & \textbf{Param} & \textbf{T-Param}
        & \multicolumn{4}{c}{\textbf{CAMO}}
        & \multicolumn{4}{c}{\textbf{CHAMELEON}}
        & \multicolumn{4}{c}{\textbf{COD10K}}
        & \multicolumn{4}{c}{\textbf{NC4K}} \\
        \cmidrule(lr){4-7}\cmidrule(lr){8-11}\cmidrule(lr){12-15}\cmidrule(lr){16-19}
        & (M) & (M)
        & MAE$\downarrow$ & $F^{\omega}\uparrow$ & $S_m\uparrow$ & $E_{\phi}\uparrow$
        & MAE$\downarrow$ & $F^{\omega}\uparrow$ & $S_m\uparrow$ & $E_{\phi}\uparrow$
        & MAE$\downarrow$ & $F^{\omega}\uparrow$ & $S_m\uparrow$ & $E_{\phi}\uparrow$
        & MAE$\downarrow$ & $F^{\omega}\uparrow$ & $S_m\uparrow$ & $E_{\phi}\uparrow$ \\
        \midrule

        \multicolumn{19}{c}{\textbf{\textit{Specialist COD Methods (CNN \& Transformer-based)}}} \\
        \midrule
        FSPNet \cite{huang2023feature} & 274.2 & 274.2 & .060 & .761 & .864 & .880 & .036 & .766 & .887 & .889 & .034 & .671 & .843 & .858 & .043 & .772 & .878 & .891\\
        ZoomNet \cite{pang2022zoom} & 33.4 & 33.4 & .070 & .731 & .809 & .856 & .030 & .812 & .881 & .928 & .031 & .719 & .833 & .882 & .045 & .774 & .847 & .887 \\
        DGNet \cite{ji2023deep} & 8.30 & 8.30 & .057 & .769 & .839 & .901 & .029 & .816 & .890 & .934 & .033 & .693 & .822 & .877 & .042 & .784 & .857 & .907 \\
        CamoFormer \cite{yin2024camoformer} & 71.3 & 71.3 & .048 & .824 & .868 & .923 & .022 & .860 & .901 & .942 & .023 & .777 & .868 & .928 & .031 & .841 & .888 & .935 \\
        ZoomNeXt \cite{pang2024zoomnext} & 28.46 & 28.46 & .041 & .857 & .892 & .935 & .020 & .865 & .903 & .941 & .018 & .834 & .898 & .939 & .030 & .859 & .902 & .932 \\

        \midrule
        \multicolumn{19}{c}{\textbf{\textit{SAM-based Adaptation Methods}}} \\
        \midrule
        \rowcolor{gray!5}
        SAM (Oracle, 1-pt GT)$^*$ & 641.08 & 0 & .201 & .460 & .580 & .574 & .221 & .484 & .588 & .582 & .083 & .583 & .710 & .714 & .135 & .579 & .678 & .683 \\
        \rowcolor{gray!5}
        SAM2 (Oracle, 1-pt GT)$^*$ & 224.51 & 0 & .086 & .714 & .771 & .802 & .041 & .814 & .848 & .888 & .030 & .794 & .850 & .888 & .046 & .811 & .849 & .883 \\
        SAM \cite{kirillov2023segment} & 641.08 & 0 & .251 & .105 & .413 & .329 & .196 & .077 & .433 & .326 & .157 & .056 & .449 & .344  & .239 & .096 & .426 & .340 \\
        SAM2 \cite{ravi2024sam} & 224.51 & 0 & .198 & .348 & .658 & .506 & .182 & .311 & .660 & .512 & .132 & .287 & .669 & .531 & .159 & .393 & .722 & .572 \\
        YOLOv8+SAM2(Auto) & 268.1 & 0 & .219 & .134 & .450 & .372 & .464 & .154 & .422 & .323 & .318 & .112 & .463 & .400 & .387 & .175 & .461 & .375 \\
        SAM2-AMG \cite{ravi2024sam} & 224.51 & 0 & .329 & .115 & .358 & .443 & .500 & .113 & .283 & .327 & .262 & .151 & .424 & .522 & .302 & .207 & .423 & .507 \\
        SAM-Adapter \cite{chen2023sam} & 641.27 & 4.25 & .072 & .763 & .836 & .858 & .034 & .801 & .878 & .891 & .025 & .797 & .876 & .904 & .043 & .810 & .875 & .891 \\
        MedSAM \cite{ma2024segment} & 93.74 & 93.73 & .069 & .757 & .847 & .873 & .036 & .800 & .895 & .918 & .037 & .718 & .854 & .892 & .047 & .791 & .878 & .905 \\
        MDSAM \cite{gao2024multi} & 101.06 & 14.57 & .061 & .782 & .839 & .885 & .027 & .838 & .891 & .942 & .030 & .752 & .846 & .904 & .042 & .811 & .866 & .908 \\
        SAM2-Adapter \cite{chen2024sam2} & 224.51 & 3.94 & .048 & .817 & .884 & .915 & .025 & .847 & .909 & .945 & .020 & .814 & .891 & .936 & .030 & .851 & .906 & .934 \\
        SAM2(SAM2-L, r80) \cite{chen2024sam2} & 234.79 & 18.96 & .037 & .862 & .900 & .935 & .022 & .857 & .908 & .942 & .020 & .817 & .890 & .933 & .029 & .860 & .904 & .936 \\
        \midrule
        \multicolumn{19}{c}{\textbf{\textit{IP-SAM Variants (Ours)}}} \\
        \midrule
        IP-SAM (SAM2-L, FT)  & 231.86 & 231.86 & .040 & .862 & .900 & .933 & .017 & .894 & .929 & .969 & .018 & .834 & .901 & .942 & .030 & .858 & .901 & .935 \\
        IP-SAM (SAM2-T, r32)   & 45.53  & 16.92  & .148 & .415 & .632 & .649 & .097 & .477 & .705 & .728 & .080 & .407 & .674 & .711 & .105 & .520 & .711 & .729 \\
        IP-SAM (SAM2-B, r32)  & 89.07  & 18.07  & .093 & .644 & .770 & .795 & .049 & .708 & .835 & .876 & .045 & .626 & .790 & .839 & .060 & .721 & .826 & .858 \\
        IP-SAM (SAM2-L, r16)  & 234.47 & 18.65  & .040 & .851 & .896 & .929 & .022 & .850 & .907 & .944 & .020 & .821 & .894 & .939 & .029 & .863 & .907 & .941 \\
        \rowcolor{gray!15}
        \textbf{IP-SAM (SAM2-L, r32)} & \textbf{237.08} & \textbf{21.26}
        & \textbf{.032} & \textbf{.878} & \textbf{.912} & \textbf{.946}
        & \textbf{.018} & \textbf{.874} & \textbf{.920} & \textbf{.956}
        & \textbf{.017} & \textbf{.838} & \textbf{.906} & \textbf{.947}
        & \textbf{.026} & \textbf{.873} & \textbf{.911} & \textbf{.943} \\
        IP-SAM (SAM2-L, r64)  & 242.30 & 26.48  & .037 & .867 & .904 & .938 & .019 & .874 & .919 & .956 & .017 & .841 & .904 & .947 & .027 & .871 & .912 & .943 \\
        \bottomrule
    \end{tabular}
    }
    
\noindent{\raggedright \scriptsize \textbf{Notes.} Param/T-Param in millions (M). $^*$Oracle uses a ground-truth centroid point to show theoretical capacity. IP-SAM metrics are averaged over 3 seeds. \textbf{FT}: Full fine-tuning. \textbf{T/B/L}: Hiera variants. \textbf{AMG}: Auto Mask Generator (top-1 via predicted IoU).\par}
\end{table}

\noindent \textbf{Discussion \& Generalization.}
Compared to specialist models, IP-SAM consistently reduces errors without human prompts. To ascertain native capacity, we introduce a zero-shot Oracle baseline (Table~\ref{tab:sota}) explicitly prompted with a ground-truth centroid. Interestingly, our automatic IP-SAM outperforms the zero-shot SAM2-Large Oracle (MAE 0.017 vs. 0.030 on COD10K). While the Oracle lacks task-specific adaptation, this reveals a crucial insight: in severe camouflage, single points are highly ambiguous, causing native decoders to segment sub-parts or leak. By synthesizing dense intrinsic prompts with gating, IP-SAM resolves these semantic conflicts far beyond point-prompting limits. Validating intrinsic conditioning over explicit auto-prompting, a zero-shot YOLOv8+SAM2 baseline frequently collapses (35.4\% full-image fallbacks on COD10K), misleading SAM2 to segment backgrounds (MAE 0.318). Against controlled feature-space adaptations, IP-SAM shows larger gains on structure-sensitive metrics ($S_m$, $E_{\phi}$), supporting deceptive cue neutralization prior to decoding. Crucially, IP-SAM maintains strong zero-shot generalization on CHAMELEON and NC4K. Efficiency-wise, IP-SAM runs at 29 FPS (vs. 35 FPS for base SAM2), introducing minimal computational overhead. Notably, compared to the feature-adapted SAM2 (SAM2-L, r80), which updates the original SAM2 decoder, IP-SAM achieves substantially superior disambiguation. Furthermore, our baseline (Table~\ref{tab:ab_core}, row 1) confirms that tuning the decoder without intrinsic prompts yields suboptimal results (MAE 0.024), proving our SOTA gains stem fundamentally from the prompt-space paradigm rather than merely unfreezing the decoder. Finally, confirming statistical robustness, IP-SAM achieves a stable MAE of $0.017 \pm 0.001$ across three random seeds on COD10K.

\subsection{Ablation Study}
\label{sec:ablation}

\noindent \textbf{Setup and core components.}
We systematically evaluate each module at $512\times512$. Our baseline (Tab.~\ref{tab:ab_core}, row 1) trains only the image-encoder LoRA ($r=32$) and our task-specific mask decoder (T-Param=9.15M in total), utilizing a frozen prompt encoder with null prompts (empty sparse tokens and a zero-mask input) for determinism. Tab.~\ref{tab:ab_core} supports our architectural evolution: (i) integrating SPG yields an immediate performance leap, suggesting that synthesizing dense self-prompts effectively reactivates the decoding pathway; (ii) introducing PSG yields the most pronounced error reduction on CAMO (dropping MAE to 0.032), indicating that asymmetric suppression is crucial under severe camouflage; (iii) lateral inhibition provides auxiliary structural regularization; (iv) the refinement head sharpens boundary fidelity orthogonally to prompt-level disambiguation.

\begin{table}[tb] 
  \centering
  \caption{\textbf{Ablation on core components.} Incrementally adding SPG, PSG, Lateral, and Refine to the baseline.}
  \label{tab:ab_core}
  \setlength{\tabcolsep}{3pt}
  \renewcommand{\arraystretch}{0.9} 
  \scriptsize
  \resizebox{\linewidth}{!}{
  \begin{tabular}{c c c c c cccc cccc}
    \toprule
    \multirow{2}{*}{SPG} & \multirow{2}{*}{PSG} & \multirow{2}{*}{Lat.} & \multirow{2}{*}{Ref.} &
    \multirow{2}{*}{T-Param} &
    \multicolumn{4}{c}{\textbf{COD10K}} &
    \multicolumn{4}{c}{\textbf{CAMO}} \\
    \cmidrule(lr){6-9}\cmidrule(lr){10-13}
    & & & & (M) &
    MAE$\downarrow$ & $F^{\omega}\uparrow$ & $S_m\uparrow$ & $E_{\phi}\uparrow$ &
    MAE$\downarrow$ & $F^{\omega}\uparrow$ & $S_m\uparrow$ & $E_{\phi}\uparrow$ \\
    \midrule
    \xmark & \xmark & \xmark & \xmark & 9.15  &
    0.024 & 0.788 & 0.857 & 0.902 &
    0.043 & 0.818 & 0.859 & 0.893 \\
    \cmark & \xmark & \xmark & \xmark & 17.00 &
    0.022 & 0.802 & 0.872 & 0.916 &
    0.041 & 0.826 & 0.875 & 0.915 \\
    \cmark & \cmark & \xmark & \xmark & 18.98 &
    0.019 & 0.826 & 0.893 & 0.938 &
    0.034 & 0.868 & 0.902 & 0.938 \\
    \cmark & \cmark & \cmark & \xmark & 19.13 &
    0.018 & 0.835 & 0.899 & 0.941 &
    0.032 & 0.871 & 0.907 & 0.942 \\
    \rowcolor{gray!15}
    \cmark & \cmark & \cmark & \cmark & \textbf{21.26} &
    \textbf{0.017} & \textbf{0.838} & \textbf{0.906} & \textbf{0.947} &
    \textbf{0.032} & \textbf{0.878} & \textbf{0.912} & \textbf{0.946} \\
    \bottomrule
  \end{tabular}
  }
  
\noindent{\raggedright \scriptsize \textbf{Notes.} Baseline (row 1) trains only the image-encoder LoRA and our task-specific mask decoder with null prompts. \cmark~indicates the module is enabled.\par}
\end{table}

\begin{table}[!b] 
  \centering
\caption{\textbf{Ablation on design choices.} (I) Conditioning paradigm and SPG capacity, (II) PSG operator and residual anchoring, (III) PSG and token replacement, and (IV) LoRA injection strategy.}
  \label{tab:ab_design}
  \renewcommand{\arraystretch}{1} 
  \setlength{\tabcolsep}{2.5pt} 
  \scriptsize 
  \resizebox{\linewidth}{!}{
  \begin{tabular}{l c cccc cccc}
    \toprule
    \multirow{2}{*}{\textbf{Setting}} & \multirow{2}{*}{\textbf{T-Param (M)}} &
    \multicolumn{4}{c}{\textbf{COD10K}} & \multicolumn{4}{c}{\textbf{CAMO}} \\
    \cmidrule(lr){3-6}\cmidrule(lr){7-10}
    & &
    MAE$\downarrow$ & $F^{\omega}\uparrow$ & $S_m\uparrow$ & $E_{\phi}\uparrow$ &
    MAE$\downarrow$ & $F^{\omega}\uparrow$ & $S_m\uparrow$ & $E_{\phi}\uparrow$ \\
    
    \midrule

    \multicolumn{10}{l}{\textbf{(I) Conditioning paradigm \& SPG Capacity}} \\
    \midrule
    Minimal-SPG (Direct) & 13.85 &
    0.051 & 0.802 & 0.887 & 0.896 & 0.058 & 0.727 & 0.873 & 0.885 \\
    Minimal-SPG + PromptSpace & 13.85 &
    0.040 & 0.851 & 0.894 & 0.928 & 0.049 & 0.818 & 0.893 & 0.936 \\
    Full-SPG (Direct) & 21.26  &
    0.025 & 0.765 & 0.883 & 0.899 &
    0.043 & 0.829 & 0.897 & 0.913 \\
    FeatAdd & 19.02 &
    0.020 & 0.825 & 0.896 & 0.940 &
    0.039 & 0.864 & 0.904 & 0.936 \\
    Trainable Prompt Encoder & 21.26 & 
    0.019 & 0.820 & 0.891 & 0.938 & 
    0.039 & 0.854 & 0.895 & 0.932 \\ 
    \rowcolor{gray!15}
    \textbf{Full-SPG + PromptSpace (Ours)} & \textbf{21.26} &
    \textbf{0.017} & \textbf{0.838} & \textbf{0.906} & \textbf{0.947} &
    \textbf{0.032} & \textbf{0.878} & \textbf{0.912} & \textbf{0.946} \\
    \midrule
    
    \multicolumn{10}{l}{\textbf{(II) PSG operator and residual anchoring}} \\
    \midrule
    None (no interaction) & 19.27 &
    0.020 & 0.818 & 0.877 & 0.918 &
    0.040 & 0.830 & 0.881 & 0.910 \\
    Subtraction ($Z^+\!-\!Z^-$) & 19.93 &
    0.019 & 0.829 & 0.898 & 0.941 &
    0.038 & 0.863 & 0.902 & 0.936 \\
    Concatenation & 20.37 &
    0.019 & 0.823 & 0.893 & 0.934 &
    0.038 & 0.851 & 0.895 & 0.933 \\
    CrossAttn Fusion & 19.53 & 
    0.019 & 0.828 & 0.899 & 0.942 &
    0.037 & 0.867 & 0.904 & 0.938 \\
    Anchor to $\widetilde{Z}^+$ & 21.26 & 
    0.019 & 0.820 & 0.891 & 0.935 &
    0.039 & 0.852 & 0.894 & 0.927 \\
    \rowcolor{gray!15}
    \textbf{AsymGate (Anchor $Z^+$)} & \textbf{21.26} &
    \textbf{0.017} & \textbf{0.838} & \textbf{0.906} & \textbf{0.947} &
    \textbf{0.032} & \textbf{0.878} & \textbf{0.912} & \textbf{0.946} \\
    \midrule

    \multicolumn{10}{l}{\textbf{(III) PSG and token replacement}} \\ 
    \midrule
    w/o PSG (Refine kept) & 19.11 &
    0.021 & 0.816 & 0.887 & 0.931 &
     0.040 & 0.846 & 0.891 & 0.922 \\
    w/ PSG + default mask tokens & 21.26 &
    0.019 & 0.828 & 0.898 & 0.940 &
    0.039 & 0.859 & 0.900 & 0.933 \\
    \rowcolor{gray!15}
    \textbf{Full IP-SAM} & \textbf{21.26} & 
    \textbf{0.017} & \textbf{0.838} & \textbf{0.906} & \textbf{0.947} &
    \textbf{0.032} & \textbf{0.878} & \textbf{0.912} & \textbf{0.946} \\
    \midrule

    \multicolumn{10}{l}{\textbf{(IV) LoRA injection strategy}} \\
    \midrule
    $qkv$ only & 19.51 &
    0.019 & 0.821 & 0.895 & 0.937 &
    0.038 & 0.859 & 0.900 & 0.935 \\
    Deep stages & 20.87 &
    0.034 & 0.689 & 0.822 & 0.873 &
    0.066 & 0.734 & 0.825 & 0.853 \\
    Shallow stages & 16.42 &
    0.024 & 0.772 & 0.869 & 0.915 &
    0.057 & 0.785 & 0.857 & 0.886 \\
    Sparse (50\%) & 18.63 &
    0.019 & 0.820 & 0.894 & 0.937 &
    0.039 & 0.859 & 0.899 & 0.932 \\
    \rowcolor{gray!15}
    \textbf{Full (ours)} & \textbf{21.26} &
    \textbf{0.017} & \textbf{0.838} & \textbf{0.906} & \textbf{0.947} &
    \textbf{0.032} & \textbf{0.878} & \textbf{0.912} & \textbf{0.946} \\
    \bottomrule
  \end{tabular}
  }
  
\noindent{\raggedright \scriptsize 
\textbf{Notes.} \textit{Direct} denotes directly using the SPG positive logit as the final mask, bypassing the prompt-conditioned decoding pipeline (i.e., prompt encoder, PSG, and mask decoder). 
\texttt{FeatAdd} adds dense prompts to image embeddings. 
\texttt{None} uses only $Z^+$, discarding $Z^-$. 
Detailed settings follow Sec.~\ref{sec:ablation}. \par}
\end{table}

\noindent \textbf{Key design choices.}
Tab.~\ref{tab:ab_design} supports our core design philosophy. \textbf{(I) Conditioning paradigm \& SPG Capacity:} To definitively isolate the contribution of Prompt-Space Conditioning (PSC) from the SPG's parameter capacity, we introduce a Minimal-SPG baseline (replacing the TwoWayTransformer with a 3-layer CNN). Stripped of heavy parameters, the Minimal-SPG's direct output yields a severely degraded MAE of 0.051 on COD10K. Crucially, projecting these coarse logits through our prompt-space paradigm and the prompt-conditioned decoding pipeline slashes the MAE to 0.040. This large error reduction suggests that reactivating the native prompt-conditioned decoding pathway plays a critical role in improving performance. Scaling back up to the Full-SPG yields our SOTA MAE of 0.017 (vs. its direct output MAE of 0.025). This confirms a clear division of labor: while a high-capacity SPG is necessary to extract reliable regional anchors in cluttered scenes, SOTA fine-grained disambiguation is ultimately unlocked by the prompt-conditioned decoder operating in the frozen prompt space. Furthermore, conditioning strictly through the native manifold (PromptSpace) consistently outshines conventional feature-space injection (FeatAdd), and unfreezing the prompt encoder degrades performance, validating that keeping it strictly frozen preserves geometric priors against camouflage noise. \quad \textbf{(II) PSG internal operator:} We compare our asymmetric gate against standard feature fusion techniques. Asymmetric gating outperforms symmetric concatenation and subtraction, which tend to equally amplify background and foreground signals in highly deceptive scenes. Furthermore, anchoring the residual to the purified $\widetilde{Z}^+$ degrades performance compared to anchoring to the original $Z^+$ (AsymGate). This validates our hypothesis regarding residual learning in prompt space: the asymmetric gate inherently risks attenuating delicate foregrounds alongside background noise. By anchoring the residual to the uncorrupted $Z^+$, we force fusion $\psi$ to act as a dedicated noise cancellation predictor rather than a feature reconstructor. This simplifies optimization, better preserving ultra-fine boundaries. This destructive interference is visually corroborated by feature energy adjustments (\figref{fig:vis_psg}, Col 8), where the network generates precise negative activations to neutralize hallucinated noise. \quad \textbf{(III) PSG vs. Token Replacement:} Removing PSG (while retaining refinement) degrades performance, indicating it provides preemptive disambiguation that post-hoc refinement cannot replicate. We isolate dynamic token replacement (w/ PSG + default mask tokens) by withholding $\mathcal{T}_{\text{prop}}$. While $\mathcal{T}_{\text{prop}}$ adds gains, the network without it still substantially outperforms baselines, confirming PSG remains the primary driver. \quad \textbf{(IV) LoRA strategy:} Finally, distributed LoRA across all blocks (Full) yields the optimal balance compared to partial injection strategies.

\begin{figure*}[tb]
\centering
\includegraphics[width=0.93\textwidth]{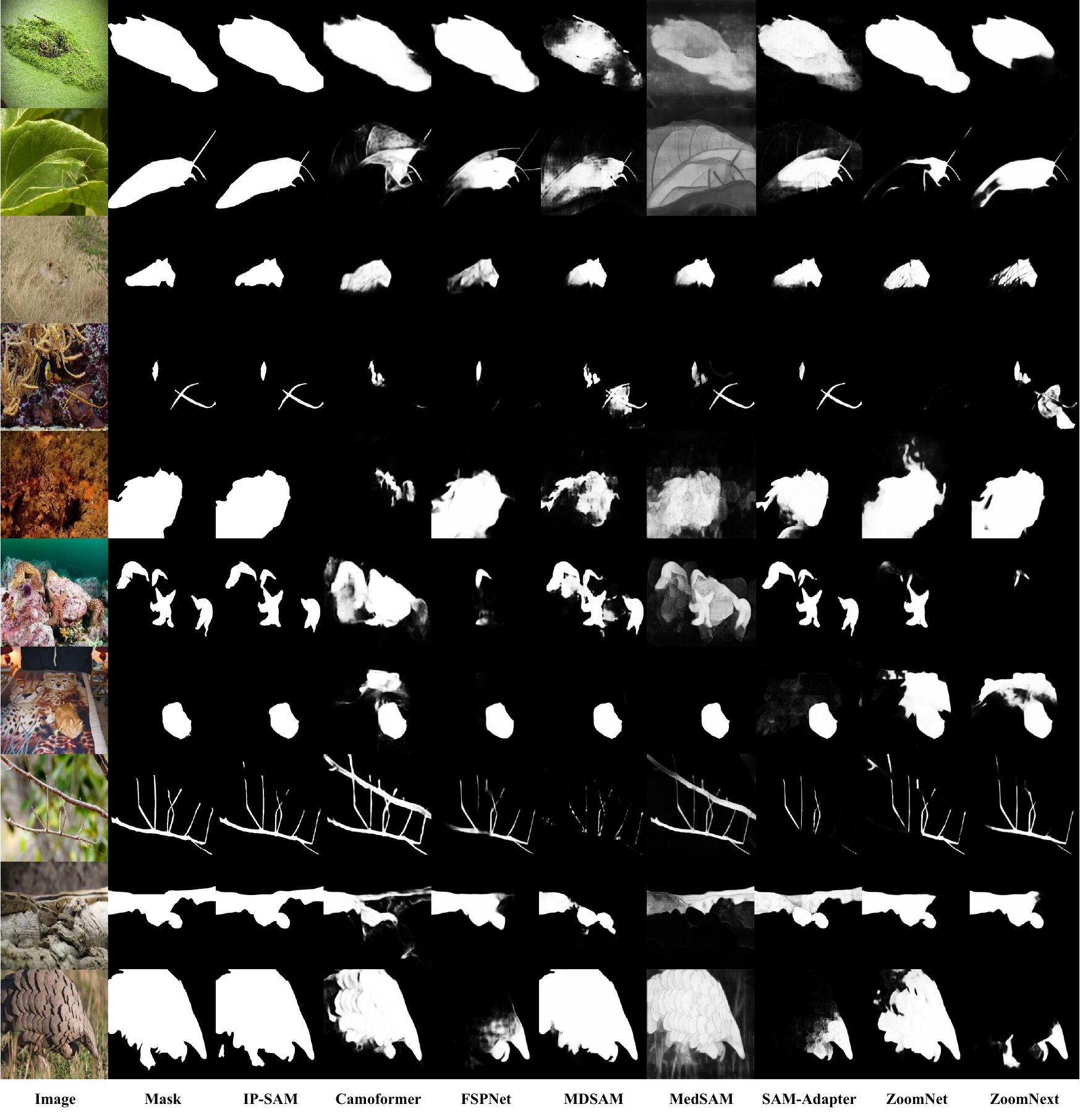}
\caption{\textbf{Qualitative comparison on challenging camouflaged scenarios.} Visual predictions of our IP-SAM against representative specialist COD models and SAM-based adaptations under the prompt-absent setting.}
\label{fig:qual}
\end{figure*}

\subsection{Qualitative Results}
\label{subsec:qual}

\noindent\textbf{Robust Ambiguity Resolution.}
As illustrated in \figref{fig:qual}, the dominant failure mode in prompt-absent COD is the hallucination of background-induced false positives alongside eroded target boundaries. Deprived of explicit navigational cues, baseline adaptations frequently struggle to decouple targets from homologous textured clutter. By contrast, IP-SAM yields substantially cleaner predictions. By utilizing intrinsic background prompts as asymmetric suppressive constraints, our PSG mechanism preemptively neutralizes deceptive cues at their representational source. Concurrently, strict alignment with SAM2's native prompt embedding space allows IP-SAM to fully exploit SAM2's pre-trained geometric priors, better preserving delicate, low-contrast topologies (\eg, thin insect antennas) even under extreme visual ambiguity.

\begin{figure*}[t]
\centering
\includegraphics[width=1\textwidth]{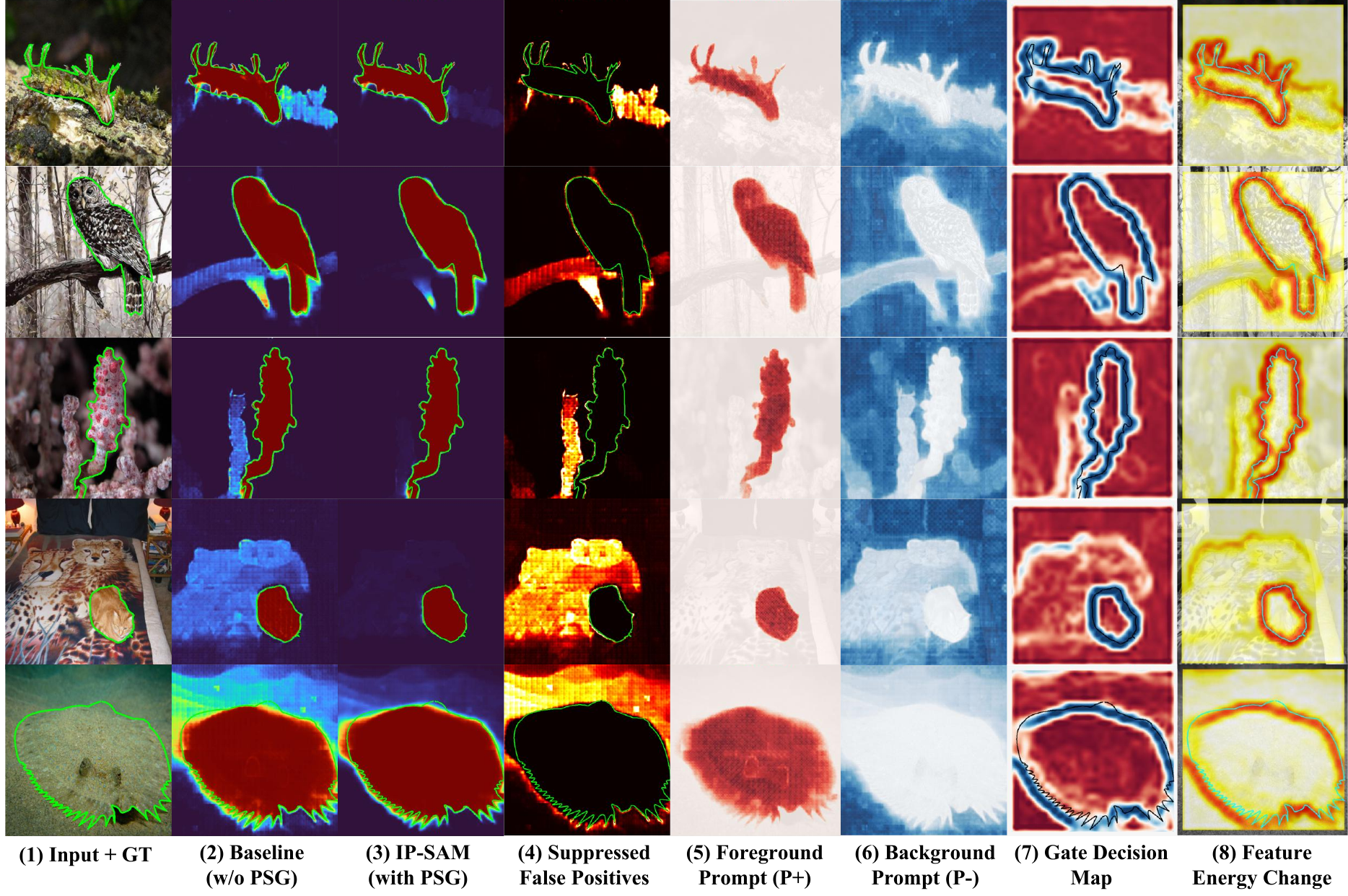}
\caption{\textbf{Internal feature visualization of the Prompt-Space Gating (PSG) mechanism.} Feature evolution from the baseline's leakage (Col 2) to IP-SAM's clean prediction (Col 3). Intermediate columns depict suppressed false positives (Col 4), complementary intrinsic prompts (Cols 5-6), gate decision map (Col 7), and feature energy adjustments (Col 8).}
\label{fig:vis_psg}
\end{figure*}

\noindent\textbf{Visualizing the Prompt-Space Gating Mechanism.}
To demystify the internal workings of our prompt-space conditioning, we visualize the feature evolution within the PSG module in \figref{fig:vis_psg}. 
Without explicit prompts, the baseline model (\figref{fig:vis_psg}, Col 2) suffers from severe feature spillover, erroneously activating homologous background regions. 
By introducing the Self-Prompt Generator (SPG), IP-SAM successfully disentangles the ambiguous context into two highly complementary representations: the foreground prompt ($P^+$, Col 5) anchoring the primary target, and the background prompt ($P^-$, Col 6) capturing deceptive distractors. 
Most importantly, the Gate Decision Map (Col 7) illustrates the effect of our asymmetric gating. Specifically, red regions indicate areas unaffected by the suppression, while dark blue regions represent the suppressive constraints derived explicitly from the encoded background prompt $Z^-$ (from $P^-$). Instead of symmetrically fusing $P^+$ and $P^-$, the network actively leverages $Z^-$ to construct a precise suppressive boundary (the dark blue "isolation ring" in Col 7) that typically envelopes the camouflaged target. 
This interpretable boundary precisely neutralizes deceptive leakage prior to entering the decoder, corroborated by the feature energy change map (Col 8, where warmer colors indicate larger energy adjustments). Consequently, IP-SAM effectively eliminates severe background-induced false positives (Col 4), ensuring that pure, target-specific prompt embeddings guide the final segmentation (\figref{fig:vis_psg}, Col 3).

\subsection{Cross-Dataset Generalization}

\begin{table}[t] 
  \centering
  \caption{\textbf{Cross-dataset medical polyp segmentation.}
  Models are trained on Kvasir-SEG and evaluated on CVC-ClinicDB and ETIS.
  Best overall are underlined and best SAM-based are boldfaced (IP-SAM belongs to both categories).}
  \label{tab:medical_transfer}
  \renewcommand{\arraystretch}{0.95}
  \setlength{\tabcolsep}{1.5pt} 
  \scriptsize
  \resizebox{\linewidth}{!}{
  \begin{tabular}{l cccc cccc cccc}
    \toprule
    \multirow{2}{*}{\textbf{Method}} & 
    \multicolumn{4}{c}{\textbf{Kvasir-SEG} (In-domain)} & 
    \multicolumn{4}{c}{\textbf{CVC-ClinicDB}} &
    \multicolumn{4}{c}{\textbf{ETIS}} \\
    \cmidrule(lr){2-5}\cmidrule(lr){6-9}\cmidrule(lr){10-13}
    & mDice$\uparrow$ & mIoU$\uparrow$ & $S_m\uparrow$ & MAE$\downarrow$ 
    & mDice$\uparrow$ & mIoU$\uparrow$ & $S_m\uparrow$ & MAE$\downarrow$
    & mDice$\uparrow$ & mIoU$\uparrow$ & $S_m\uparrow$ & MAE$\downarrow$ \\
    \midrule

    SEPNet \cite{wang2024polyp} 
    & .896 & .839 & .917 & .028 
    & .815 & .744 & .874 & .028
    & .741 & .663 & .847 & .018 \\
    LACFormer \cite{van2023lacformer} 
    & .906 & .851 & .922 & .029
    & .826 & .758 & .889 & \underline{.026}
    & \underline{.798} & \underline{.722} & \underline{.884} & \underline{.016} \\
    ASPS \cite{li2024asps} 
    & .788 & .704 & .844 & .063
    & .670 & .580 & .786 & .069
    & .362 & .313 & .660 & .039 \\
    \midrule
    ZoomNeXt \cite{pang2024zoomnext} 
    & .892 & .831 & .909 & .031 
    & .805 & .731 & .863 & .031
    & .693 & .625 & .841 & .023 \\
    SAM-Adapter \cite{chen2023sam} 
    & .624 & .514 & .724 & .117 
    & .450 & .366 & .671 & .093
    & .223 & .189 & .596 & .070 \\
    MDSAM \cite{gao2024multi} 
    & .906 & .851 & .922 & .028 
    & .796 & .724 & .863 & .034
    & .753 & .692 & .858 & .021 \\
    SAM2-Adapter \cite{chen2024sam2} 
    & .899 & .846 & .918 & .031 
    & .801 & .738 & .873 & .030
    & .733 & .663 & .847 & .026 \\
    \rowcolor{gray!15}
    \textbf{IP-SAM (Ours)} 
    & \underline{\textbf{.913}} & \underline{\textbf{.861}} & \underline{\textbf{.929}} & \underline{\textbf{.025}} 
    & \underline{\textbf{.840}} & \underline{\textbf{.778}} & \underline{\textbf{.897}} & \textbf{.029}
    & \textbf{.764} & \textbf{.696} & \textbf{.868} & \textbf{.021} \\
    \bottomrule
  \end{tabular}
  }

\noindent{\raggedright \scriptsize 
\textbf{Notes.} All SAM-based adaptations and specialist models were re-trained by us on the Kvasir-SEG training split and evaluated under the identical strictly prompt-absent protocol.\par}
\end{table}

\noindent\textbf{Cross-Dataset Medical Segmentation.}
To demonstrate the generality of our prompt-space conditioning paradigm beyond camouflaged object detection, we evaluate IP-SAM on medical polyp segmentation under the same prompt-absent protocol. The model is trained solely on the Kvasir-SEG training split using the same hyperparameter configuration as in the COD experiments, without introducing domain-specific augmentation or task-specific heuristics.

As shown in Table~\ref{tab:medical_transfer}, IP-SAM achieves the best in-domain performance on Kvasir-SEG and consistently outperforms both specialist models and representative SAM-based adaptations (e.g., MDSAM and SAM2-Adapter), all of which were re-trained under the same training protocol for fair comparison. On CVC-ClinicDB, IP-SAM achieves the highest mDice (0.840) while remaining competitive in MAE, matching or surpassing representative specialist models such as LACFormer. On the more challenging ETIS dataset, which exhibits notable domain shifts due to varying endoscopic equipment and smaller polyp scales, IP-SAM also achieves the best performance among all SAM-based baselines.

These results suggest that conditioning through SAM2's frozen prompt interface provides a stable adaptation pathway for prompt-absent deployment while mitigating the overfitting risks associated with injecting task-specific signals directly into intermediate features. This suggests that intrinsic prompt synthesis may better exploit the foundation model’s pre-trained geometric priors, rather than relying solely on task-specific feature modifications.

\section{Conclusion and Limitations}
\label{sec:conclusion}

We address the structural paradox of deploying prompt-conditioned segmenters in prompt-absent scenarios, where unguided decoders tend to hallucinate false positives. Rather than bypassing the native prompt interface through feature-space patching, IP-SAM adapts SAM2-based segmentation through prompt-space conditioning: SPG synthesizes intrinsic regional anchors encoded by the frozen prompt encoder, while PSG suppresses background-induced interference prior to decoder interaction. Under a strict prompt-absent protocol on COD benchmarks, IP-SAM establishes a new state-of-the-art with only 21.26M trainable parameters. The same paradigm also demonstrates strong cross-dataset generalization on medical polyp segmentation.

Despite these advantages, IP-SAM has limitations. First, its effectiveness depends on the capacity of the underlying foundation model; performance decreases when scaling down to smaller variants such as SAM2-T/B. Second, in extremely low-contrast scenarios where SPG fails to separate foreground and background ($P^-$ may capture the target), asymmetric gating may over-suppress target signals. Although the residual anchoring ($+Z^+$) partially alleviates this issue by preserving the original prompt embedding, future work will explore uncertainty-aware prompt generation to better handle such extreme cases.

\section{Acknowledgments}
This study was supported in part by National Natural Science Foundation of China under Grants 61971404 and 61901454; in part by the Project of Technology and Engineering Center for Space Utilization, Chinese Academy of Sciences under Grant T503471; in part by the Youth Innovation Promotion Association of the Chinese Academy of Sciences under Grant 2019168; and in part by the Project of Technology and Engineering Center for Space Utilization, Chinese Academy of Sciences under Grant CSU-JJKT-2023-4.

\newpage
\appendix
\onecolumn
\noindent{\Large\bfseries Appendix}\par
\vspace{0.5em}

This appendix is organized to directly address potential review risks associated with the core claims of the main paper:
\begin{itemize}[leftmargin=*, topsep=2pt, itemsep=4pt]
    \item \textbf{Claim 1: The proposed prompt-space conditioning paradigm transfers beyond camouflage-heavy settings.}\\
    \textit{Review risk:} The method may be too specialized to COD.\\
    \textit{Evidence provided:} Sections \ref{sec:supp_metrics} and \ref{sec:sod_qualitative} report quantitative and qualitative results on five standard Salient Object Detection (SOD) benchmarks using the exact same architecture and recipe.
    
    \item \textbf{Claim 2: The method exhibits qualitatively stable behavior across ambiguous and cross-domain scenarios.}\\
    \textit{Review risk:} Performance might degrade under representative distractors or when transferred to visually distinct domains.\\
    \textit{Evidence provided:} Sections \ref{sec:polyp_qualitative} and \ref{sec:more_cod_results} provide qualitative evidence across medical polyps and extreme camouflaged targets.
    
    \item \textbf{Claim 3: The primary gain stems from prompt-space conditioning, not merely decoder redesign.}\\
    \textit{Review risk:} Improvements could be attributed to the stronger task-specific decoder or extra parameters.\\
    \textit{Evidence provided:} Section \ref{sec:ablation_prompt_space} decouples conditioning pathways and logit formats under an identical MaskDecoderV2 and training recipe to isolate the source of gains.
    
    \item \textbf{Claim 4: The framework remains computationally practical.}\\
    \textit{Review risk:} The prompt-space modules may introduce substantial hidden runtime overhead.\\
    \textit{Evidence provided:} Section \ref{sec:efficiency_analysis} details both wall-clock latency and synchronized component-wise profiling.
    
    \item \textbf{Claim 5: The observed failure modes are bounded and admit a consistent qualitative explanation.}\\
    \textit{Review risk:} The gating mechanism might fail unpredictably under severe ambiguity.\\
    \textit{Evidence provided:} Section \ref{sec:failure_cases} visualizes the coupled failure process and relates it to upstream feature ambiguity.
\end{itemize}

\setcounter{section}{0}
\renewcommand{\thesection}{\Alph{section}}
\renewcommand{\thesubsection}{\thesection.\arabic{subsection}}

\section{Generalization on Salient Object Detection}
\label{sec:supp_metrics}

To further evaluate whether our prompt-space conditioning paradigm generalizes beyond camouflage-heavy and medical settings, we test IP-SAM on five standard RGB Salient Object Detection (SOD) benchmarks: DUTS-TE \cite{wang2017learning}, DUT-OMRON \cite{yang2013saliency}, HKU-IS \cite{li2015visual}, ECSSD \cite{yan2013hierarchical}, and PASCAL-S \cite{li2014secrets}. This supplementary evaluation is intended to explicitly test whether the advantage of conditioning through SAM2's native prompt interface is specific to extreme aliasing, or if it transfers effectively to standard foreground-background segmentation tasks. 

\begin{table}[!tb]
\vspace{-4mm}
  \centering
  \caption{\textbf{Quantitative comparison on standard RGB Salient Object Detection (SOD) benchmarks.} Results of competing methods are quoted from their original papers when available. ``-'' indicates the metric is not reported. $F_\beta^{\max}$ and $E_m^{\max}$ denote the maximum F-measure and maximum E-measure, respectively. The best results are highlighted in \textbf{bold}.}
  \label{tab:sod_generalization}
  \renewcommand{\arraystretch}{1.1} 
  \setlength{\tabcolsep}{3.5pt} 
  \scriptsize
  \resizebox{\linewidth}{!}{
  \begin{tabular}{l | cccc | cccc | cccc | cccc | cccc}
    \toprule
    \multirow{2}{*}{\textbf{Method}} & 
    \multicolumn{4}{c|}{\textbf{DUTS-TE}} & 
    \multicolumn{4}{c|}{\textbf{DUT-OMRON}} & 
    \multicolumn{4}{c|}{\textbf{HKU-IS}} & 
    \multicolumn{4}{c|}{\textbf{ECSSD}} & 
    \multicolumn{4}{c}{\textbf{PASCAL-S}} \\
    \cmidrule(lr){2-5}\cmidrule(lr){6-9}\cmidrule(lr){10-13}\cmidrule(lr){14-17}\cmidrule(lr){18-21}
    & MAE$\downarrow$ & $F_\beta^{\max}\uparrow$ & $S_m\uparrow$ & $E_m^{\max}\uparrow$ 
    & MAE$\downarrow$ & $F_\beta^{\max}\uparrow$ & $S_m\uparrow$ & $E_m^{\max}\uparrow$ 
    & MAE$\downarrow$ & $F_\beta^{\max}\uparrow$ & $S_m\uparrow$ & $E_m^{\max}\uparrow$ 
    & MAE$\downarrow$ & $F_\beta^{\max}\uparrow$ & $S_m\uparrow$ & $E_m^{\max}\uparrow$ 
    & MAE$\downarrow$ & $F_\beta^{\max}\uparrow$ & $S_m\uparrow$ & $E_m^{\max}\uparrow$ \\
    \midrule

    \multicolumn{21}{c}{\textbf{\textit{CNN-Based Methods}}} \\
    \midrule
    F3Net \cite{wei2020f3net}
    & .035 & .905 & .888 & .920 
    & .053 & .841 & .838 & .864 
    & .028 & .943 & .917 & .952 
    & .033 & .957 & .924 & .948 
    & .061 & .892 & .861 & .898 \\

    GateNet-X \cite{zhao2020suppress}
    & .035 & .908 & .897 & .916 
    & .051 & .847 & .849 & .865 
    & .029 & .946 & .925 & .947 
    & .035 & .957 & .929 & .944 
    & .064 & .892 & .865 & .895 \\

    MINet-R \cite{pang2020multi} 
    & .037 & .884 & .884 & .917 
    & .056 & .831 & .833 & .860 
    & .029 & .942 & .919 & .952 
    & .033 & .954 & .925 & .950 
    & .064 & .881 & .856 & .896 \\
    
    MENet \cite{wang2023pixels} 
    & .028 & .918 & .905 & .938 
    & .045 & .845 & .850 & .871 
    & .023 & .951 & .927 & .960 
    & .021 & .957 & .928 & .951 
    & .053 & .897 & .872 & .910 \\

    \midrule
    \multicolumn{21}{c}{\textbf{\textit{Transformer-Based Methods}}} \\
    \midrule
    VST \cite{liu2021visual} 
    & .037 & .895 & .896 & .919 
    & .058 & .836 & .850 & .871 
    & .029 & .946 & .928 & .952 
    & .033 & .954 & .932 & .951 
    & .061 & .882 & .872 & .902 \\

    SelfReformer \cite{yun2022selfreformer}
    & .027 & .920 & .911 & .943 
    & .043 & .853 & .861 & .884 
    & .024 & .949 & .931 & .960 
    & .027 & .959 & .936 & .957 
    & .051 & .902 & .881 & .919 \\

    EBM \cite{zhang2021learning}
    & - & .900 & .909 & .949 
    & - & .817 & .858 & .900 
    & - & .943 & .930 & .971 
    & - & .954 & .941 & .972 
    & - & .856 & .877 & .899 \\
    
    ICON-S \cite{zhuge2022salient} 
    & .025 & .924 & .917 & .954 
    & .043 & .862 & .869 & .900 
    & .022 & .954 & .935 & .968 
    & .023 & .962 & .914 & .968 
    & .048 & .903 & .885 & .924 \\

    BBRF \cite{ma2023boosting} 
    & .025 & .911 & .909 & .949 
    & .044 & .839 & .861 & .896 
    & .020 & .949 & .932 & .969 
    & .022 & .961 & .939 & .969 
    & .049 & .887 & .878 & .923 \\

    DC-Net-S \cite{zhu2025dc} 
    & .023 & .932 & .925 & .952 
    & .039 & .868 & .875 & .898 
    & .021 & .957 & .941 & .966 
    & .023 & .968 & .947 & .965 
    & .049 & .904 & .887 & .917 \\

    VSCode-S \cite{luo2024vscode} 
    & - & .922 & .926 & .960 
    & - & .840 & .877 & .912 
    & - & .951 & .940 & .974 
    & - & .959 & .949 & .974 
    & - & .864 & .887 & .904 \\

    Samba \cite{he2025samba} 
    & - & .930 & .932 & \textbf{.966} 
    & - & .859 & .889 & \textbf{.922} 
    & - & .956 & .945 & \textbf{.978} 
    & - & .965 & .953 & \textbf{.978} 
    & - & .896 & .892 & .931 \\

    \midrule
    \multicolumn{21}{c}{\textbf{\textit{Foundation Model Adaptations (SAM-based)}}} \\
    \midrule
    SAM-Adapter \cite{chen2023sam} 
    & .050 & .814 & .824 & .821 
    & .051 & .770 & .804 & .795 
    & .041 & .908 & .886 & .899 
    & .047 & .919 & .902 & .908 
    & .087 & .818 & .817 & .826 \\

    SAM2-Adapter \cite{chen2024sam2} 
    & .022 & .934 & .935 & .956 
    & .041 & .841 & .881 & .905 
    & .019 & .954 & .944 & .968 
    & .020 & .963 & .951 & .966 
    & .043 & .893 & .897 & .928 \\

    SAM2-UNet \cite{xiong2026sam2} 
    & .020 & - & .934 & .959 
    & .039 & - & .884 & .912 
    & .019 & - & .941 & .971 
    & .020 & - & .950 & .970 
    & .043 & - & .894 & .931 \\

    SAM2-LPNet \cite{zhu2025sam2} 
    & .029 & .891 & .877 & -
    & .040 & .840 & .850 & - 
    & .024 & .938 & .918 & - 
    & .025 & .950 & .930 & - 
    & .052 & .889 & .864 & - \\
    
    MDSAM \cite{gao2024multi} 
    & .024 & .937 & .920 & .949 
    & .039 & \textbf{.887} & .878 & .910 
    & .019 & \textbf{.963} & .941 & .969 
    & .021 & \textbf{.974} & .948 & .967 
    & .052 & \textbf{.907} & .882 & .917 \\
    
    \rowcolor{gray!15}
    \textbf{IP-SAM (Ours)} 
    & \textbf{.019} & \textbf{.940} & \textbf{.941} & .963 
    & \textbf{.034} & .863 & \textbf{.896} & .919 
    & \textbf{.017} & .960 & \textbf{.950} & .973 
    & \textbf{.018} & .965 & \textbf{.955} & .970 
    & \textbf{.041} & .900 & \textbf{.899} & \textbf{.934} \\
    \bottomrule
  \end{tabular}
  }
\end{table}

\begin{figure}[!tb]
  \vspace{-2mm} 
  \centering
  \includegraphics[width=1\textwidth]{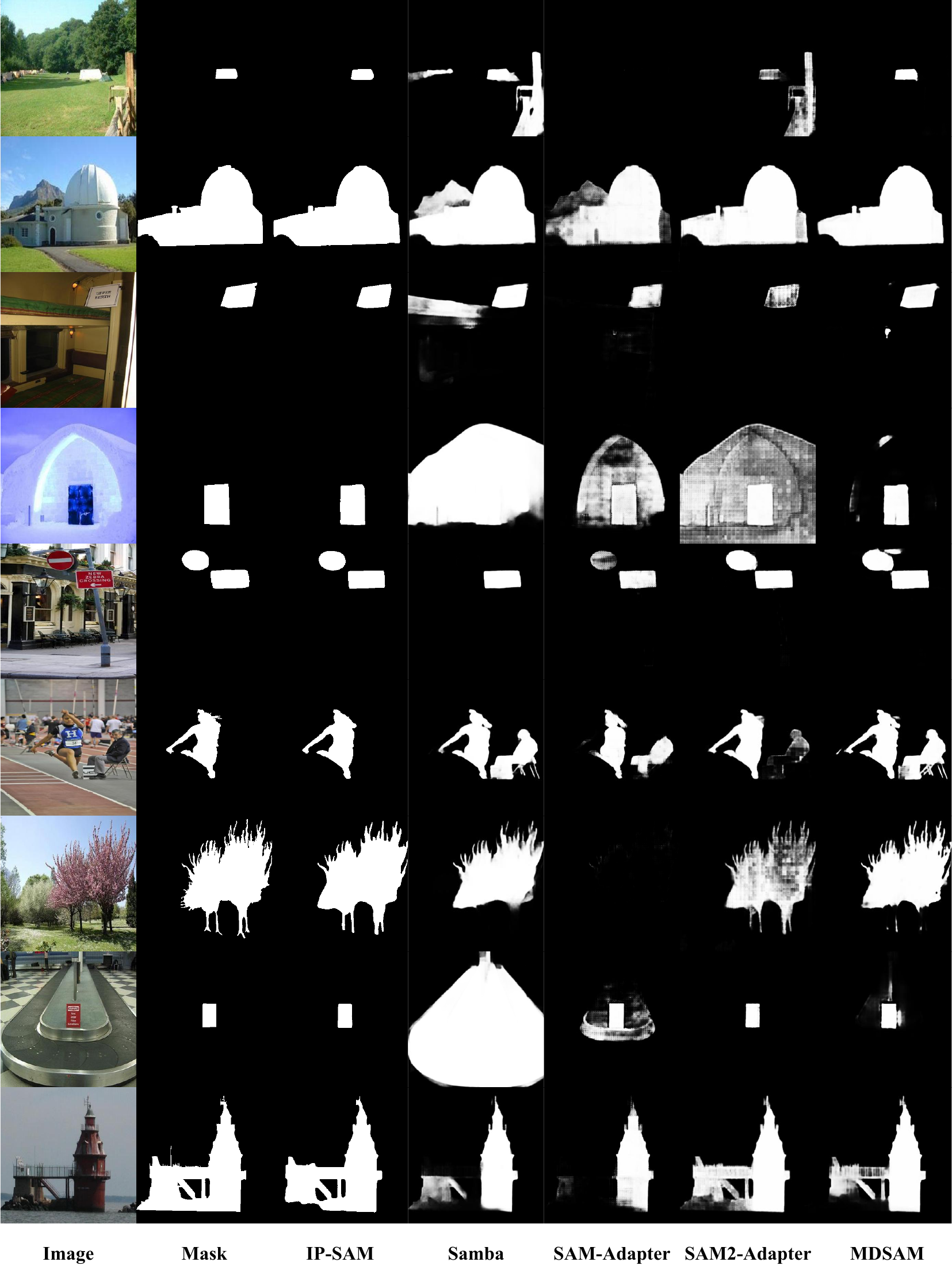} 
  \vspace{-5mm} 
  \caption{\textbf{Qualitative comparison on Salient Object Detection (SOD).} Visual results demonstrate IP-SAM's precise localization and resistance to over-segmentation. A detailed analysis is provided in Section \ref{sec:sod_qualitative}.}
  \label{fig:sod_success}
\end{figure}

Following the standard RGB SOD protocol, our model is trained on the DUTS-TR training split. Crucially, we maintain the exact same network architecture and training recipe as in our main experiments, without introducing any task-specific architectural modifications.

As detailed in Table \ref{tab:sod_generalization}, we compare IP-SAM against a broad spectrum of 14 state-of-the-art models to assess whether the proposed prompt-space conditioning transfers beyond the main COD setting. This supplementary study is intended primarily to evaluate transferability under a fixed adaptation recipe, rather than to establish an exhaustive retraining benchmark on SOD. Under this setting, IP-SAM achieves the best MAE and Structure-measure ($S_m$) across all five datasets, while remaining highly competitive on $F_\beta^{\max}$ and $E_m^{\max}$.

\section{Qualitative Analysis on Salient Object Detection}
\label{sec:sod_qualitative}

The qualitative examples in Figure \ref{fig:sod_success} are intended as visual support for the quantitative SOD results in Section \ref{sec:supp_metrics}. They show that IP-SAM remains selective in scenes with strong structural distractors (e.g., the igloo in Row 4 and the baggage carousel in Row 8) without notable background leakage. Furthermore, it successfully preserves multi-part structures (e.g., the street sign and lighthouse scaffolding) with accurate boundaries. We view these observations as consistent with our broader generalization claim, demonstrating resistance to over-segmentation rather than serving as an independent proof of mechanism.

\begin{figure}[!tb]
  \vspace{-2mm}
  \centering
  \includegraphics[width=1\textwidth]{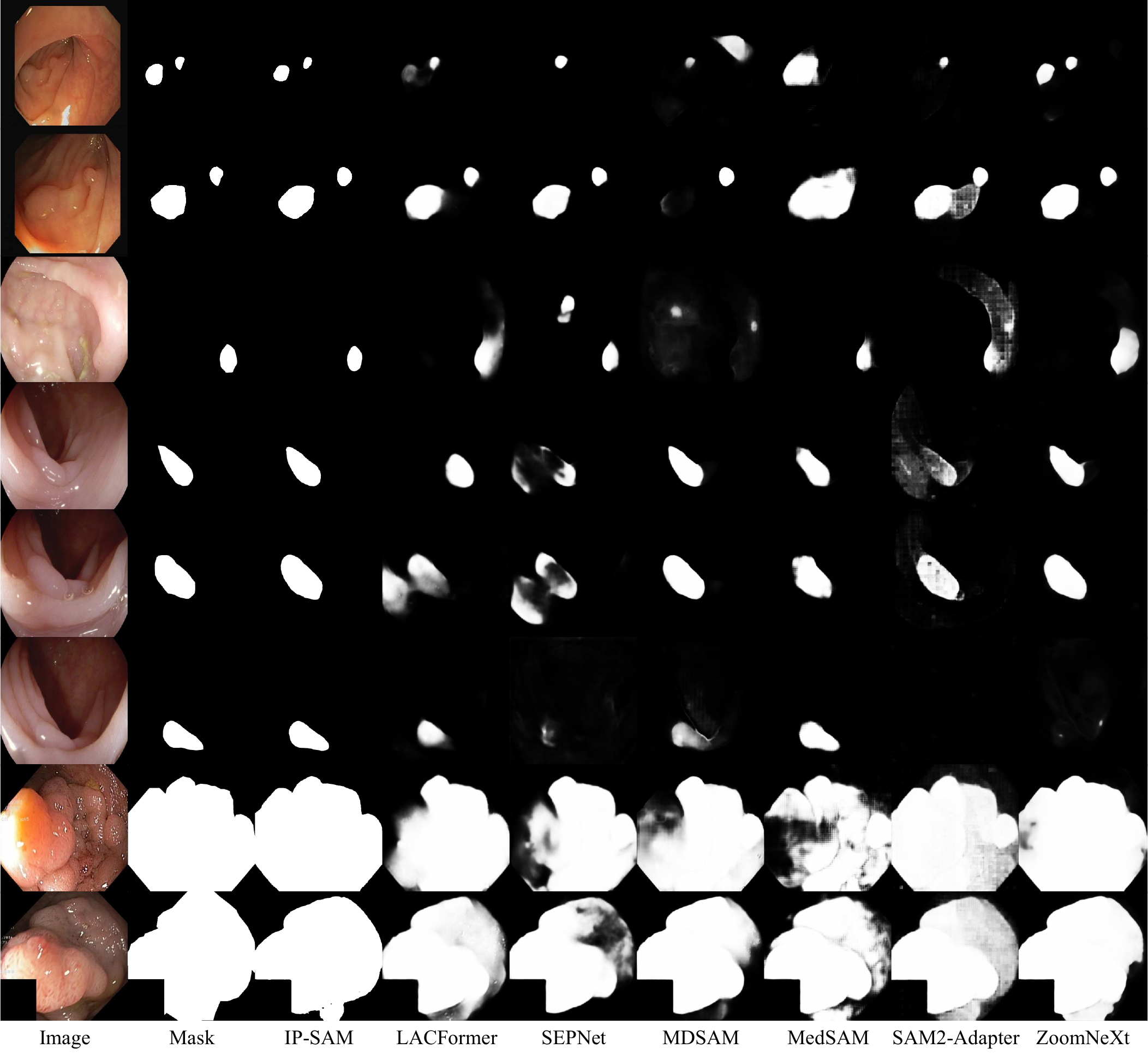} 
  \vspace{-5mm}
  \caption{\textbf{Qualitative comparison on medical polyp segmentation.} IP-SAM demonstrates robust boundary preservation and effectively mitigates mucosa-induced false positives. See Section \ref{sec:polyp_qualitative} for detailed observations.}
  \label{fig:polyp_vis}
\end{figure}

\section{Qualitative Analysis on Medical Polyp Segmentation}
\label{sec:polyp_qualitative} 

To visually contextualize the cross-dataset transferability reported in the main paper, Figure \ref{fig:polyp_vis} presents typical challenging scenarios in medical polyp segmentation, including extreme scale variations and low-contrast mucosal boundaries. While baseline adaptations often struggle with under-segmentation on small polyps or background leakage on low-contrast edges, IP-SAM exhibits relatively stable performance. It effectively mitigates background-induced false positives and preserves structural completeness. These observations align with the quantitative transferability results and are consistent with the view that prompt-space suppressive constraints can help reduce ambiguous context in this setting.

\section{Additional Qualitative Results on Camouflaged Object Detection}
\label{sec:more_cod_results}

\begin{figure}[!tb]
  \vspace{-2mm}
  \centering
  \includegraphics[width=1\textwidth]{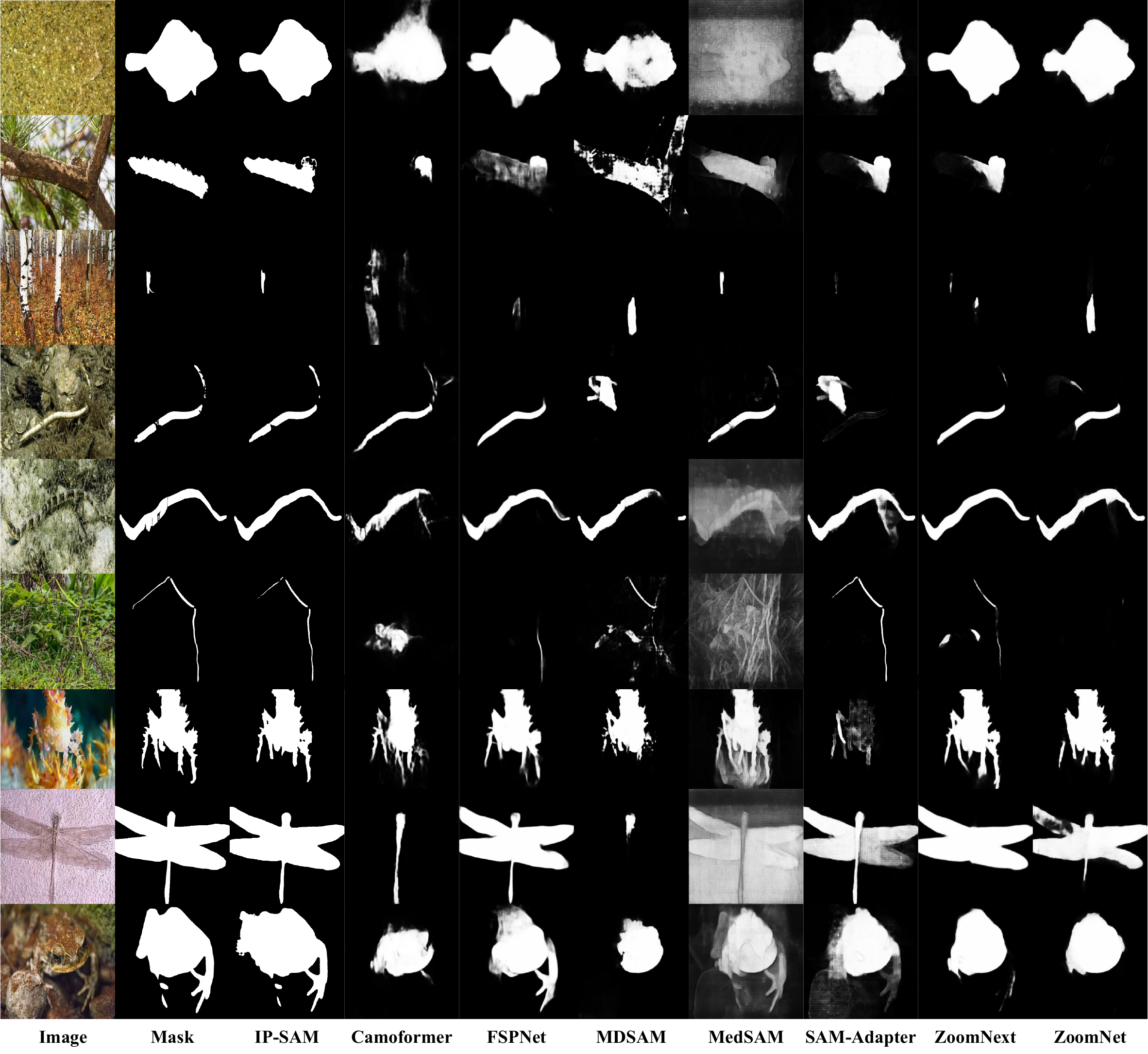} 
  \vspace{-3mm}
  \caption{\textbf{Additional qualitative results on extreme camouflaged scenarios.} Visual comparison of IP-SAM against representative COD models and SAM adaptations under the prompt-absent setting. IP-SAM demonstrates robust disambiguation across near-zero contrast (top rows), deceptive distractors (middle rows), and fragile topologies (bottom rows).}
  \label{fig:more_cod_success}
\end{figure}

As supplementary visual evidence to the main text, Figure \ref{fig:more_cod_success} illustrates representative success patterns of IP-SAM under extreme Camouflaged Object Detection (COD) scenarios. These examples span near-zero contrast (top rows), deceptive distractors (middle rows), and fragile topologies (bottom rows). In these challenging cases, IP-SAM generally yields cleaner masks and better preserves fine, continuous structures compared to feature-space adaptations. Together with the failure boundary analysis in Section \ref{sec:failure_cases}, these examples provide a more complete picture of the model's operating characteristics under severe visual ambiguity.

\section{Ablation Study on Prompt-Space Conditioning}
\label{sec:ablation_prompt_space}

To rigorously validate the core mechanisms of our prompt-space conditioning paradigm, we conduct a decoupled ablation study on the prompt conditioning pathways and the input logit formats, as detailed in Table \ref{tab:ab_design}. To ensure absolute fairness and strictly isolate the source of performance gains, all variants in this study share the identical downstream task-specific decoder architecture (MaskDecoderV2), initialized with the exact same weights and trained under identical recipes.

\noindent\textbf{Importance of Routing Through the Frozen Prompt Encoder.} 
A natural question arises: Does the performance improvement simply stem from introducing a new task-specific decoder? To address this, we evaluate three conditioning pathways in Table \ref{tab:ab_design} (I). 
First, No Prompt Conditioning completely disables the Self-Prompt Generator (SPG)---deliberately reducing trainable parameters to 13.40M to strictly isolate the module's effect---and feeds zero dense embeddings to the decoder, which yields suboptimal results (e.g., MAE of 0.020 on COD10K). 
Next, we introduce the Bypass Prompt Encoder variant, where the SPG generates intrinsic logits, but these are directly mapped to dense embeddings via a lightweight convolutional projection ($\sim$34K params), actively bypassing SAM2's frozen prompt encoder. Notably, this feature-space shortcut yields negligible improvement over the null prompt baseline. 
In stark contrast, routing the intrinsic prompts through the frozen prompt encoder (Prompt-Space Cond.) drops the MAE to 0.017 and significantly boosts $F^{\omega}$ and $S_m$. These results support the view that the performance gain is closely tied to routing intrinsic signals through the frozen prompt encoder, beyond decoder redesign alone.

\noindent\textbf{Takeaway: The gain does not stem from decoder replacement alone.} Under an identical MaskDecoderV2 and training recipe, substantial improvement is observed only when intrinsic prompts are routed through SAM2’s frozen prompt encoder, whereas bypassing the prompt encoder or using the decoder alone provides negligible benefit.

\begin{table}[!t] 
  \centering
  \caption{\textbf{Ablation study on prompt conditioning pathways and input logit formats.} All variants share an identical downstream MaskDecoderV2 architecture to strictly isolate the source of performance gains. The results indicate that the observed gain is closely tied to routing intrinsic prompts through SAM2's frozen prompt encoder, especially when using raw continuous logits, rather than to decoder replacement alone.}
  \label{tab:ab_design}
  \renewcommand{\arraystretch}{0.9} 
  \setlength{\tabcolsep}{2.5pt} 
  \scriptsize 
  \resizebox{\linewidth}{!}{
  \begin{tabular}{l c cccc cccc}
    \toprule
    \multirow{2}{*}{\textbf{Setting}} & \multirow{2}{*}{\textbf{T-Param (M)}} &
    \multicolumn{4}{c}{\textbf{COD10K}} & \multicolumn{4}{c}{\textbf{CAMO}} \\
    \cmidrule(lr){3-6}\cmidrule(lr){7-10}
    & &
    MAE$\downarrow$ & $F^{\omega}\uparrow$ & $S_m\uparrow$ & $E_{\phi}\uparrow$ &
    MAE$\downarrow$ & $F^{\omega}\uparrow$ & $S_m\uparrow$ & $E_{\phi}\uparrow$ \\
    
    \midrule

    \multicolumn{10}{l}{\textbf{(I) Prompt Conditioning Pathway}} \\
    \midrule
    No Prompt Conditioning & 13.40 &
    0.020 & 0.811 & 0.890 & 0.934 & 0.040 & 0.849 & 0.896 & 0.930 \\
    Bypass Prompt Encoder & 21.26 &
    0.020 & 0.817 & 0.893 & 0.935 & 0.040 & 0.850 & 0.897 & 0.928 \\
    \rowcolor{gray!15}
    \textbf{Prompt-Space Cond. (Ours)} & \textbf{21.26} &
    \textbf{0.017} & \textbf{0.838} & \textbf{0.906} & \textbf{0.947} &
    \textbf{0.032} & \textbf{0.878} & \textbf{0.912} & \textbf{0.946} \\

    \midrule
    \multicolumn{10}{l}{\textbf{(II) Input Logit Formats for Prompt Encoder}} \\
    \midrule
    Sigmoid Probabilities & 21.26 &
    0.019 & 0.819 & 0.895 & 0.937 & 0.039 & 0.856 & 0.901 & 0.932 \\
    Hard Binarization & 21.26 &
    0.019 & 0.823 & 0.898 & 0.938 & 0.041 & 0.853 & 0.901 & 0.931 \\
    Sharpened Logits ($\times 5.0$) & 21.26 &
    0.022 & 0.789 & 0.873 & 0.925 & 0.039 & 0.855 & 0.894 & 0.935 \\
    Flattened Logits ($\times 0.2$) & 21.26 &
    0.020 & 0.814 & 0.894 & 0.936 & 0.038 & 0.856 & 0.902 & 0.934 \\
    \rowcolor{gray!15}
    \textbf{Raw Continuous Logits (Ours)} & \textbf{21.26} &
    \textbf{0.017} & \textbf{0.838} & \textbf{0.906} & \textbf{0.947} &
    \textbf{0.032} & \textbf{0.878} & \textbf{0.912} & \textbf{0.946} \\
    
    \bottomrule
  \end{tabular}
  }
  \vspace{1pt}
\noindent{\raggedright \scriptsize 
\textbf{Notes.} The lower T-Param in No Prompt reflects the disabled SPG. Bypass replaces the frozen prompt manifold with a $\sim$34K-param projection. Hard Binarization thresholds at logit=0. \par}
\end{table}

\vspace{1mm}
\noindent\textbf{Compatibility of Raw Continuous Logits.} 
Having established the necessity of the prompt encoder, we further investigate the optimal format of the dense prompts before they are ingested by the frozen encoder. Conventional practices often apply sigmoid activations or hard thresholds to mask logits. As shown in Table \ref{tab:ab_design} (II), Hard Binarization (thresholded at logit=0, equivalent to a 0.5 probability) heavily degrades performance, as converting logits to strict $\{0.0, 1.0\}$ values irreparably destroys continuous boundary details and spatial uncertainty. 
Similarly, applying Sigmoid Probabilities or intentionally altering the logit contrast (Sharpened and Flattened logits) leads to suboptimal metrics. We attribute this to the fact that artificial scaling disrupts the natural input distribution expected by the pre-trained frozen encoder. 
Conversely, directly feeding Raw Continuous Logits yields the best performance across all metrics. These comparisons are consistent with the view that raw continuous logits gracefully preserve fine-grained spatial uncertainty while remaining better matched to the input characteristics expected by the frozen prompt encoder.

\noindent \textbf{Takeaway.} The prompt encoder is most effective when receiving raw continuous intrinsic logits, suggesting that preserving their original spatial uncertainty is more advantageous than post-hoc thresholding or rescaling.

\begin{table}[!t]
  \centering
  \caption{\textbf{Computational efficiency analysis of IP-SAM.} \textbf{(a)} End-to-end inference speed measured with the same wall-clock protocol as the main paper. \textbf{(b)} Internal component-wise profiling measured with synchronized CUDA Events (averaged over 200 runs after 50 warm-up iterations), used solely to illustrate the relative computational distribution inside the IP-SAM forward pass. Most computation remains in the SAM2 backbone, while the proposed prompt-space modules contribute only a modest fraction of the internal runtime.}
  \label{tab:runtime_breakdown}
  \renewcommand{\arraystretch}{1.1} 
  \setlength{\tabcolsep}{8pt} 
  \scriptsize 
  
  \begin{tabular}{l c c}
    \toprule
    \multicolumn{3}{l}{\textbf{(a) End-to-end Inference Speed (Wall-clock Protocol)}} \\
    \midrule
    \textbf{Method} & \textbf{Latency (ms)} & \textbf{FPS} \\
    \midrule
    SAM2-L (zero-prompt baseline) & $\sim$28.57 & 35 \\
    IP-SAM-L (Ours)               & $\sim$34.48 & 29 \\
    \rowcolor{gray!15}
    \textbf{Extra Overhead}       & \textbf{+5.91} & \textbf{+20.7\%} \\
    \bottomrule
  \end{tabular}

  \vspace{4mm}
  
  \begin{tabular}{l r r}
    \toprule
    \multicolumn{3}{l}{\textbf{(b) Component-wise Runtime Distribution (CUDA Events)}} \\
    \midrule
    \textbf{Internal Component} & \textbf{CUDA Latency (ms)} & \textbf{Share (\%)} \\
    \midrule
    Image Encoder (Backbone + FPN Neck) & 45.34 $\pm$ 0.22 & 72.4\% \\
    SPG (PPG + NPG Prompt Generators)   &  6.00 $\pm$ 0.03 &  9.6\% \\
    Frozen Prompt Encoder (pos + neg)   &  0.99 $\pm$ 0.01 &  1.6\% \\
    PSG / PIM Fusion                    &  1.19 $\pm$ 0.01 &  1.9\% \\
    Task Decoder (MaskDecoderV2 + PDI + PRM) &  9.12 $\pm$ 0.05 & 14.5\% \\
    \midrule
    \textbf{Total (Internal CUDA Profiling)} & \textbf{62.64} & \textbf{100.0\%} \\
    \bottomrule
  \end{tabular}
\end{table}

\section{Computational Efficiency Analysis}
\label{sec:efficiency_analysis}

In the main paper, we reported the end-to-end inference speed of IP-SAM under the same wall-clock protocol adopted by prior works. To further clarify its computational characteristics, we provide a two-level runtime analysis in Table \ref{tab:runtime_breakdown}. All measurements are conducted on a single NVIDIA RTX 3090 GPU with a batch size of 1 and an input resolution of $512 \times 512$.

For strict consistency with the main text, \textbf{Table \ref{tab:runtime_breakdown}(a)} reports the end-to-end inference speed under the standard wall-clock protocol. IP-SAM operates at 29 FPS, compared with 35 FPS for the zero-prompt SAM2-L baseline. This corresponds to a modest end-to-end latency increase of approximately 20.7\%, which preserves practical inference efficiency for prompt-absent automatic segmentation.

To transparently dissect where this overhead originates, \textbf{Table \ref{tab:runtime_breakdown}(b)} presents an internal component-wise profiling measured using strict CUDA Events synchronization. Because wall-clock timing without explicit synchronization underestimates actual GPU execution time, the latencies in Table \ref{tab:runtime_breakdown}(b) are used solely to reveal the relative internal computational distribution and are not meant to be directly summed or matched to the end-to-end numbers in Table \ref{tab:runtime_breakdown}(a). 

The breakdown explicitly reveals that the vast majority of the computation (72.4\%) remains within the SAM2 image encoder. Crucially, our proposed prompt-space modules---comprising the SPG, the frozen prompt encoder, and the PSG fusion---together account for only a modest 13.1\% of the internal runtime. The remaining overhead is mainly associated with the more expressive task-specific decoder (MaskDecoderV2 + PDI + PRM) designed for fine-grained COD prediction.

\section{Mechanistic Analysis of Failure Cases}
\label{sec:failure_cases}

\begin{figure}[!b]
  \vspace{-2mm}
  \centering
  \includegraphics[width=1\textwidth]{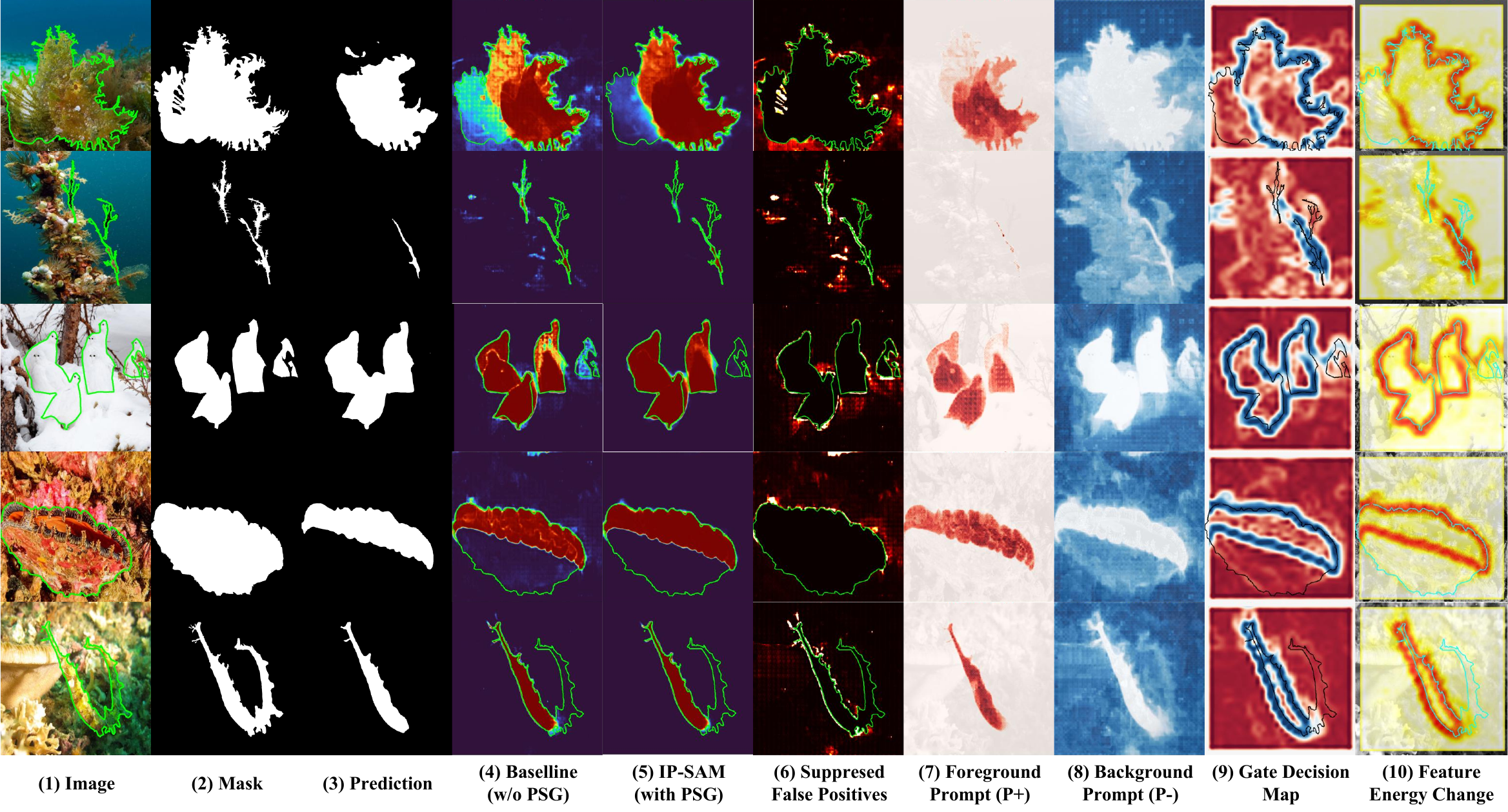} 
  \vspace{-3mm}
  \caption{\textbf{Mechanistic visualization of IP-SAM's failure modes.} By unboxing the internal feature evolution, we illustrate how extreme visual ambiguity triggers a coupled error mechanism: the negative prompt ($P^-$, Column 8) erroneously encroaches upon the target regions, establishing an aggressive suppressive gate, which further diminishes the already weakened intrinsic foreground prompt ($P^+$, Column 7). This propagates through the Prompt-Space Gating (PSG) module, ultimately resulting in severe under-segmentation.}
  \label{fig:failure_mechanisms}
\end{figure}

To better characterize the model's operating boundary, we visualize the internal feature evolution of typical failure cases in Figure \ref{fig:failure_mechanisms}. Consistent with the limitations discussed in Section 5 of the main paper, severe over-suppression (under-segmentation) remains the dominant failure mode.

In extreme scenarios where the target shares virtually identical micro-textures, lighting, or topologies with the background, the image encoder struggles to provide cleanly separable features. A close examination of the internal feature maps reveals how this limitation unfolds within our architecture: it is a compounding effect driven primarily by negative prompt encroachment ($P^-$), which actively suppresses an already weakened foreground discovery ($P^+$).

As directly observed in Column 8 ($P^-$) across all examples, the blue negative prompt not only anchors the surrounding background but actively encroaches upon the ambiguous target area. Because our PSG module is explicitly designed to leverage the negative prompt embedding as the primary asymmetric constraint, this upstream miscalibration immediately dictates the gating behavior. Consequently, the Gate Decision Map (Column 9) formulates a deep blue suppressive boundary directly over the true foreground.

Subjected to this strong suppressive prior, the red foreground prompt (Column 7, $P^+$) exhibits incomplete or highly diminished activation. Severely camouflaged, the SPG struggles to establish a comprehensive positive anchor to counteract the suppression. Facing active background gating and lacking robust positive guidance, the network is strictly instructed to filter out these features. 

This logical but erroneous propagation ultimately leads to specific structural errors (as seen in Column 3), which can be directly mapped to the visual evidence:
\begin{itemize}[leftmargin=*, topsep=2pt, itemsep=0pt]
    \item \textbf{Partial Truncation (Rows 1 and 4):} When intricate parts of the target blend deeply into the background, $P^-$ erroneously covers these areas and establishes a gating wall. Concurrently, $P^+$ remains largely inactive in these regions, causing the gate to slice off significant portions of the target.
    \item \textbf{Topological Fracture (Rows 2 and 5):} For extremely thin or fragile structures, the negative prompt encroaches into the spatial gaps. Because the $P^+$ signal is highly diminished at these thinnest points and cannot overcome the suppression, the gate severs the continuous topology.
    \item \textbf{Instance Missing (Row 3):} In multi-instance scenarios with homogeneous backgrounds, $P^-$ envelops the rightmost peripheral instance. Without a strong $P^+$ activation to anchor this specific bird, the suppressive gate leads to its complete omission.
\end{itemize}

\vspace{2mm}
\noindent\fbox{
\parbox{1\textwidth}{
\textbf{Failure Boundary Summary.} IP-SAM most often fails under extreme foreground-background indistinguishability, particularly when negative prompts encroach upon weak target regions. The dominant symptom is under-segmentation rather than uncontrolled background activation. These failures are therefore more consistent with upstream ambiguity and over-suppression than with a general collapse of the decoding pipeline. We include these examples to clarify the operating boundary of the method rather than to claim complete robustness under all ambiguous cases.
}}

\section{Final Remarks and Scope of Evidence}

We conclude by summarizing the evidence scope and operating boundary of this work. Rather than bypassing the native prompt interface through feature-space patching, IP-SAM shows that translating task-specific cues into SAM2’s frozen prompt manifold can serve as an effective and parameter-efficient adaptation route.

Our current empirical evidence is centered on prompt-absent foreground-background segmentation under strong visual ambiguity, including COD, SOD, and cross-dataset medical polyp segmentation. Within this scope, prompt-space conditioning consistently improves structural selectivity and reduces deceptive false positives relative to feature-space bypassing.

At the same time, the failure analysis in Section G suggests that the method remains challenged when foreground-background ambiguity becomes extreme and negative prompts encroach upon weak target regions. These observations clarify the present operating boundary of the method, while also indicating that prompt-space conditioning may provide a useful foundation for adapting prompt-dependent architectures beyond the current settings. Extending this paradigm to more complex multi-instance or broader semantic segmentation scenarios remains an important direction for future study.

\bibliographystyle{splncs04}
\bibliography{main}

\end{document}